\documentclass{article}

\usepackage[main, final]{neurips_2026}

\usepackage[utf8]{inputenc} 
\usepackage[T1]{fontenc}    
\usepackage{hyperref}       
\usepackage{url}            
\usepackage{booktabs}       
\usepackage{amsfonts}       
\usepackage{nicefrac}       
\usepackage{microtype}      
\usepackage{xcolor}         
\usepackage{microtype}
\usepackage{graphicx}
\usepackage{subcaption}
\usepackage{mathrsfs}
\usepackage{ulem}

\usepackage{multirow,array}

\usepackage{bbm}
\usepackage{algorithm}
\usepackage{algorithmicx}
\usepackage{algpseudocode}

\usepackage{amsmath}
\usepackage{amssymb}
\usepackage{mathtools}
\usepackage{amsthm}
\usepackage{wrapfig}

\theoremstyle{plain}

\theoremstyle{definition}

\theoremstyle{remark}

\newcommand{\T}{\mathcal{T}}
\newcommand{\I}{\mathbf{I}}
\DeclareMathOperator{\E}{\mathbb{E}}

\title{Multilevel and Sequential Monte Carlo for Training-Free Diffusion Guidance}

\author{Aidan Gleich \\ 
Department of Statistical Science \\ 
Duke University \\
\texttt{aidan.gleich@duke.edu} \\
\And 
Scott C. Schmidler \\ 
Departments of Statistical Science and Computer Science\\ 
Duke University}

\begin{document}
\maketitle

\begin{abstract}
  We address the problem of accurate, training-free guidance for conditional generation in trained diffusion models. Existing methods typically rely on point-estimates to approximate the posterior score, often resulting in biased approximations that fail to capture multimodality inherent to the reverse process of diffusion models. We propose a sequential Monte Carlo (SMC) framework that integrates over the full denoising distribution to construct an unbiased Monte Carlo estimator of $p_\theta(y\mid x_t)$, avoiding the bias of existing training-free guidance methods. To ensure computational tractability, we incorporate variance-reduction schemes based on MultiLevel Monte Carlo (MLMC). We evaluate on three image generation tasks. On CIFAR-10 and ImageNet classifier guidance, our approach improves accuracy by 18 and 38 points respectively over the strongest training-free baseline. On CelebA-HQ joint-attribute guidance, three existing methods (DPS, LGD, TFG) used within our framework all outperform the closest SMC-based comparator (TDS) by 25-37 points, and improve over the strongest baseline by 2-14 points.
  
\end{abstract}

\section{Introduction}
Diffusion models have emerged as a leading methodology for generative modeling \citep{sohl2015, ho2020, song2020, song2021}. They are capable of generating high-quality samples across a range of domains, including images \citep{iddpm, rombach2022}, audio \citep{kong2021}, molecules \citep{hoogeboom2022}, and text \citep{li2022diffusionlmimprovescontrollabletext}. A learned model $p_\theta(x_0)$ targets the true data distribution $p(x_0)$, enabling unconditional sampling from complex, high-dimensional distributions. 

In many applications, however, the goal is not to sample from the distribution $p(x_0)$ but instead the conditional distribution $p(x_0 \mid y)$, where $y$ is information such as an image label or molecule property. Given a likelihood function $p(y\mid x_0)$ (e.g. a classifier), this conditional distribution can be written in terms of the generative model using Bayes' rule: $p_\theta(x_0\mid y) \propto p_\theta(x_0)p(y\mid x_0)$. Efficient conditional sampling via diffusion models remains a central focus of the literature, with numerous approaches proposed in a variety of contexts.

Some approaches, particularly in image generation, include conditional information directly during training \citep{ho2022classifierfreediffusionguidance} or train auxiliary classifiers to guide the sampling process \citep{dhariwal2021}. While effective, these methods incur high computational costs during training and limit model flexibility; adapting to a new likelihood $p(y\mid x_0)$ requires retraining. 

To address this inflexibility, a separate line of work focuses on \textit{training-free guidance}. These methods combine a pre-trained unconditional model $p_\theta(x_0)$ with approximations to the posterior score $\nabla_{x_t} \log p_\theta(y\mid x_t)$. Diffusion Posterior Sampling (DPS) \citep{dps} pioneered this approach by approximating the intractable marginal likelihood $p_\theta(y\mid x_t)$ with a point estimate $p(y|\hat{x}_\theta(x_t))$, effectively collapsing the posterior to a Dirac mass at the conditional mean. Subsequent methods, including LGD \citep{lgd}, FreeDoM \citep{freedom}, and Universal Guidance \citep{universal_guidance}, refine this approach through Monte Carlo sampling over Gaussian kernels, recurrent denoising strategies, and energy-based optimization. \citet{tfg} unifies these methods into a common hyperparameter framework. However, they all rely on point approximations that fail to capture the multimodality of the posterior distribution $p_\theta(x_0 \mid x_t)$, introducing bias that accumulates over the diffusion trajectory.

The Twisted Diffusion Sampler (TDS) \citep{wuTDS} offers a theoretically distinct formulation, framing conditional sampling as Sequential Monte Carlo (SMC) with learned ``twisting functions'' to approximate the optimal proposal distribution. While this SMC perspective provides asymptotic exactness in principle, practical implementations of TDS construct their twisting functions using the same point estimate approximations discussed above. Consequently, TDS inherits the fundamental limitation of heuristic guidance: it fails to integrate over the full posterior $p_\theta(x_0 \mid x_t)$ in the proposal step, leading to weight degeneracy in multimodal settings.

To address these shortcomings, we propose an SMC framework to sample from conditional distributions using only a pre-trained unconditional model. Unlike biased heuristics that approximate the score guidance term, our approach targets the exact conditional distribution by utilizing an unbiased estimator of the marginal likelihood $p_\theta(y \mid x_t)$. To render this estimator computationally tractable, we employ a Multilevel Monte Carlo (MLMC) scheme that integrates over the full denoising distribution $p_\theta(x_0 \mid x_t)$.  Our contributions are threefold: (1) we formulate a sequential Monte Carlo framework that leverages unbiased likelihood estimators to target the exact conditional distribution, enabling the rigorous correction of biased heuristics; (2) we extend the MLMC framework of \citet{hajiali2025} to the training-free guidance problem and show empirically that it reduces the variance of likelihood estimates at a fraction of the cost of naive Monte Carlo while introducing negligible bias; and (3) we demonstrate that this approach improves the performance of existing training-free guidance methods across three conditional image generation tasks. We evaluate our approach using both accuracy and cost-per-success: the expected wall-clock cost to generate at least one valid sample with high probability. On tasks where baseline methods have low accuracy (e.g. ImageNet), our framework beats baselines on both, while on tasks with high baseline accuracy, it provides accuracy improvements at compute overhead while beating the closest SMC-based comparator.

\section{Background}

\subsection{Diffusion Models}
A diffusion model is a generative model trained to sample from an unknown distribution $p(x_0)$ represented by a finite set of samples $x_0^1,\ldots,x_0^n$. It does so by reversing a predefined forward process that iteratively noises data points via a Markov chain $q(x_t \mid x_{t-1})$. The Markov chain is 
defined such that by time $T$ the noised data approximately follows a simple distribution, 
(e.g. standard Gaussian).

In the Denoising Diffusion Probabilistic Model (DDPM) formulation \citep{ho2020}, this forward process is defined by a finite sequence of time steps $t \in \{1,\ldots,T\}$ as $q(x_t \mid x_{t-1}) = \text{N}(x_t;\sqrt{\alpha_t}x_{t-1}, \allowbreak (1-\alpha_t)\I),$ where the parameters $\{\alpha_t\}_{t=1}^T$ control the noise level of the forward process.
Noised data points can then be sampled from the forward process in a single step: 
\[q(x_t \mid x_0) = \text{N}(x_t;\sqrt{\overline{\alpha}_t}x_0,(1-\overline{\alpha}_t)\I)\]
where $\overline{\alpha_t} = \prod_{i=1}^t \alpha_i$, and these samples are used to train a neural network $\epsilon_\theta(x_t,t)$ to predict the noise component $\epsilon$ of $x_t = \sqrt{\overline{\alpha}_t}x_0 + \sqrt{1 - \overline{\alpha}_t}\epsilon$. Via Tweedie's formula, this also yields an estimate of the score of the marginal $q_t(x_t)$ of the forward process \citep{song2020}. We denote the implied score network as $s_\theta(x_t, t) := -\epsilon_\theta(x_t,t) / \sqrt{1 - \overline{\alpha}_t} \approx \nabla_{x_t}\log q(x_t).$

The network $\epsilon_\theta(x_t,t)$ is then used to define the reverse process $p_\theta(x_{t-1} \mid x_{t})$. This learned process is trained to approximate the forward process $q(x_{t-1} \mid x_t,x_0)$, which is a tractable Gaussian when conditioned on $x_0$. Typically, the reverse process is also defined as Gaussian:
\[p_\theta(x_{t-1} \mid x_t) = \text{N}(x_{t-1}; \mu_\theta(x_t,t), \sigma_t^2\I).\]

The mean $\mu_\theta(x_t,t)$ is predicted directly from $\epsilon_\theta$ via a closed-form expression \citep{ho2020}. Once $\epsilon_\theta$ is trained, sampling proceeds by drawing $x_T \sim \mathcal{N}(0,\I)$ and iterating the reverse kernels $p_\theta(x_{t-1}\mid x_t)$ for $t=T,\ldots,1$.

\subsection{Sequential Monte Carlo}
Sequential Monte Carlo (SMC) methods are a broad class of algorithms designed to sample from complex target distributions $\pi(x)$ \citep{smc_samplers}. For our purposes, SMC can be viewed as a generalization of importance sampling: the target distribution is represented by a set of $N$ weighted particles $\{x^{(i)}, w^{(i)}\}_{i=1}^N$ generated by the algorithm. The sampler is 
constructed by defining a sequence of distributions $\mu_0\ldots,\mu_{V-1}, \mu_V$ 
terminating in the target distribution ($\mu_V= \pi$),
chosen such that $\mu_0$ is easy to sample from and $\mu_v$ is ``close'' to $\mu_{v+1}$. Sampling proceeds by drawing a set of initial particles from the distribution $\mu_0$. Each step of the algorithm begins with particles $\{x_{v-1}^{(i)}\}_{i=1}^N$ distributed approximately according to $\mu_{v-1}$. The algorithm then performs three steps:
\begin{enumerate}
    \item \textbf{Propagate:} Particles are moved via a proposal distribution
    \item \textbf{Reweight:} Particles are assigned an importance weight $w_v^{(i)} = \frac{\mu_v(x_v^{(i)})}{\mu_{v-1}(x_{v-1}^{(i)}) r_v(x_v^{(i)}|x_{v-1})}$ to correct for the proposal.
    \item \textbf{Resample:} A new set $\{x_v^{(i)}\}_{i=1}^N$ is sampled with replacement from the weighted particles, yielding an unweighted approximation of $\mu_v$.
\end{enumerate}
Each step can be viewed as sampling/importance resampling (SIR) \citep{rubinSIR} with target distribution $\mu_v$. 
As $N\to\infty$, the empirical measure formed by the resampled particles provides a consistent estimator of the target distribution $\mu_v$.

\subsection{Diffusion as SMC}
As noted by \citet{wuTDS}, the reverse process of the diffusion model can be reformulated as an SMC algorithm by setting the number of SMC steps equal to the number of diffusion steps $T$ and aligning the sequence of target distributions $\{\mu_t\}_{t=T}^0$ with the marginal distributions of the reverse process $p_\theta(x_t)$ at the finite set of times $t \in \{0,\ldots,T \}$. The resulting SMC algorithm has target distributions that exactly mirror the reverse process of the diffusion model \citep{smcdiff, mcgdiff, wuTDS}. 
Sampling begins by drawing \textit{i.i.d.} particles from initial distribution $\mu_T = p(x_T) = \text{N}(x_T;0,\I)$. Proposal distributions are given by the trained reverse process: $r_t(x_{t-1}\mid x_t) = p_\theta(x_{t-1}\mid x_t)$. In the absence of conditioning information, the proposals target the desired (approximate) distribution by construction, the importance weights are uniform, and the reweighting and resampling steps of the SMC algorithm can be omitted.

\section{Conditional Diffusion Sampling Via SMC}
Given a diffusion model $p_\theta(x_0)$ estimated to approximate target $p(x_0)$ and a ``likelihood'' function $p(y \mid x_0)$ for observed information $y$ on which we wish to condition, defined over the support of $p(x_0)$, our goal is to sample from the conditional distribution $p_\theta(x_0 \mid y) \propto p_\theta(x_0)p(y \mid x_0)$ (as an approximation to $p(x_0 \mid y) \propto p(x_0)p(y \mid x_0)$) without additional training.

A naive approach generates samples from $p_\theta(x_0)$ via the unconditional reverse process, then reweights by $w_0^{(i)} \propto p(y \mid x_0^{(i)})$ and resamples. This is equivalent to SIR \citep{rubinSIR} with proposal $p_\theta(x_0)$ and target $p_\theta(x_0 \mid y)$. When these distributions differ substantially (e.g.\ in KL divergence \citep{chatterjee_diacnois_IS}), few proposed samples land in high-density regions of the target. For hard constraints $p(y\mid x_0)\in\{0,1\}$, the acceptance probability is the marginal $p_\theta(y)$, which is typically exponentially small in high dimensions.

To address this, it is preferred to choose the sequence of intermediate distributions to ensure that $\mu_t$ is close to $\mu_{t+1}$ for all $t$. 
A natural choice is to condition the reverse steps of the diffusion model, setting $\mu_t =  p_\theta(x_t \mid x_{t+1}, y)$ \citep{whiteley2014twisted, wuTDS}. If these conditional distributions can be sampled exactly then reweighting/resampling are not required, and the resulting samples are drawn exactly from the target distribution with only a single particle by the chain rule of probability. However, sampling from $p_\theta(x_t \mid x_{t+1}, y)$ requires knowledge of the intractable marginal likelihood $p_\theta(y \mid x_t)$. Accurately approximating this quantity is essential for effective conditional sampling and will be a central challenge we address.

\subsection{Conditional Reverse Steps and Marginal Likelihoods}\label{sec:marg_likes}

Many existing \textit{training-free guidance} methods \citep{dps, lgd, tfg} attempt to sample approximately from $p_\theta(x_t \mid x_{t+1}, y)$ by approximating the score of the conditional reverse process $p_\theta(x_t \mid y)$ by
\begin{align}
\nabla_{x_t} \log p_\theta(x_t \mid y) &= \nabla_{x_t}\log p_\theta(x_t) + \nabla_{x_t}\log p_\theta(y \mid x_t) \nonumber \\
&\approx s_\theta(x_t,t) + \tilde{s}_\theta(x_t,y,t)
\label{Eqn:ApproxCondScore}
\end{align}
to enable score-based guidance using the unconditional score model combined with an approximation $\tilde{s}_\theta(x_t,y,t)$ to $\nabla_{x_t}\log p_\theta(y \mid x_t)$. The difficulty is that $p_\theta(y \mid x_t)$ is intractable, requiring marginalization over $x_0,x_1,\ldots,x_{t-1}$: 
\[p_\theta(y\mid x_t) = \int p(y \mid x_0) p_\theta(x_{0:t-1} \mid x_t)dx_{0:t-1}.\]
\textit{Diffusion posterior sampling (DPS)} \citep{dps} addresses this problem by applying the approximation $\tilde{s}_\theta(x_t,y,t) = \nabla_{x_t} \log \tilde{p}(y\mid x_t)$ with
\begin{equation}
p_\theta(y\mid x_t) \approx \tilde{p}(y \mid x_t):= p_{\theta}(y \mid \hat{x}_\theta(x_t,t))
\label{Eqn:PointApprox}
\end{equation}
where $\hat{x}_\theta(x_t) \approx \E[x_0\mid x_t]$ is produced by the denoising network. This approximation replaces the full conditional distribution $p_\theta(x_0 \mid x_t)$ with a Dirac measure $\delta_{\hat{x}_\theta(x_t)}(x_0)$ at its (approximated) expected value. 
\citet{wuTDS} combine
\eqref{Eqn:ApproxCondScore}
 and \eqref{Eqn:PointApprox} to form 
 a proposal distribution for SMC.

 As noted in \citet{lgd}, this approximation can severely underestimate uncertainty in the distribution $p_\theta(x_0 \mid x_t)$. For moderate to large $t$, this distribution is almost certainly multimodal if the underlying data distribution $p(x_0)$ is; this will be the case if, for example, the unconditional target distribution $p_0(x)$ is a mixture of class-conditional densities. For multimodal distributions, the expectation is typically a poor location parameter, and the point measure
$\delta_{\hat{x}_\theta(x_t)}(x_0)$ will be a particularly poor approximation of the distribution, resulting in the targeted sequence of marginal distributions $\tilde{p}_\theta(x_t \mid y) \propto p_\theta(x_t)\tilde{p}_\theta(y\mid x_t)$ diverging from the true marginals $p_\theta(x_t\mid y)$ (see Appendix A.2). This is because the approximation shifts mass towards regions where the likelihood at the expected state $p\left(y \mid \mathbb{E}[x_0 \mid x_t]\right)$ is maximized, driving the trajectory too strongly towards the mode and ignoring uncertainty in the conditional distribution $p_\theta(x_0 \mid x_t)$.

\begin{wrapfigure}{r}{0.42\linewidth}
    \centering
    \vspace{-10pt}
    \includegraphics[width=\linewidth]{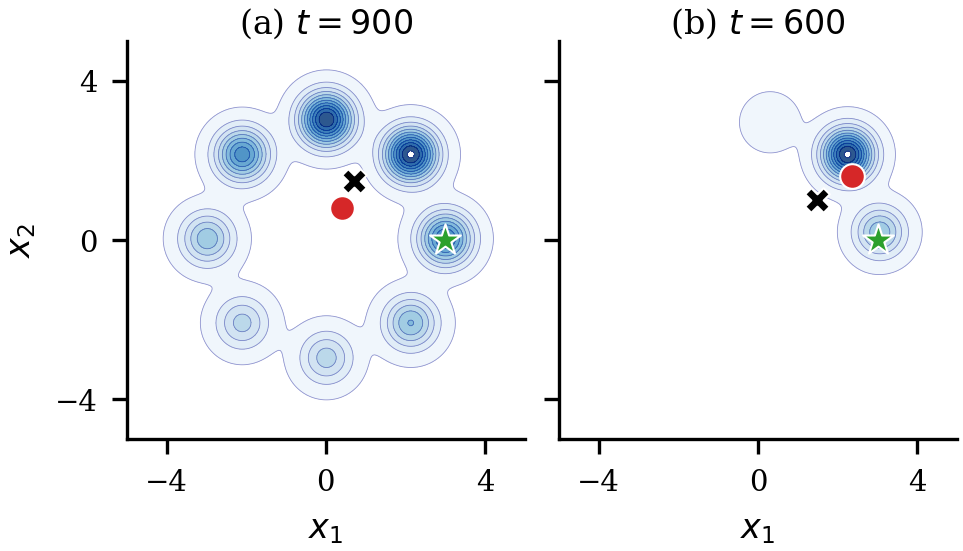}
    \caption{Posterior $p(x_0\mid x_t)$ for an 8-component ring GMM. Panels show high noise ($t=900$, left) and moderate noise ($t=600$, right).}
    \label{fig:bimod_gmm}
    \vspace{-10pt}
\end{wrapfigure}
Figure~\ref{fig:bimod_gmm} demonstrates this effect on an 8-component GMM. The noisy observation $x_t$ ({\color{black}$\times$}), the true $x_0$ ({\color{green}$\star$}), and the point estimate $\E[x_0\mid x_t]$ ({\color{red}$\bullet$}) are overlaid on contours of $p(x_0\mid x_t)$. At high noise the posterior is multimodal and the point estimate lies in a low-density region at the ring's center; at moderate noise the mass concentrates on two modes and the point estimate falls near the non-target mode. Calculating $p_\theta(y\mid x_t) = \int p(y\mid x_0)p_\theta(x_0\mid x_t)dx_0$ requires integrating over $p_\theta(x_0\mid x_t)$; replacing this distribution with a point mass at $\E[x_0\mid x_t]$ causes $p(y\mid \E[x_0\mid x_t])$ to estimate $p(y\mid x_t)$ poorly whenever the mean lands in a low-density or non-target region (see Appendix~\ref{app:gaussian_like}).

The point estimator faces further limitations when $p(y \mid x_0) \propto 1_{A_y}$ for some $A_y \subset  \mathcal{X}$ (e.g. inpainting \citep{repaint}). 
Then $p_\theta(y \mid x_t) = 
p_{\theta}(x_0 \in A_y \mid x_t)
$ is the probability 
$x_0$ 
lies in $A_y$ given $x_t$; for large $t$ this approaches the marginal: $p_\theta(x_0 \in  A_y\mid x_t) \approx p_\theta(x_0 \in  A_y)$. If the set $A_y$ has low probability under the (approximated) data distribution
$p_\theta(x_0)$, the mean estimate $\hat{x}_\theta(x_t)$ is unlikely to lie within $A_y$ for large $t$, causing $p(A_y\mid \hat{x}_\theta(x_t))$ to incorrectly estimate the probability as $0$.

\subsection{Monte Carlo Approximation to Marginal Likelihoods}\label{sec:mc_approx}
To overcome the limitations of DPS-style methods illustrated above, we instead estimate $p_\theta(y \mid x_t)$ directly by Monte Carlo 
integration. A simple MC approximation can be obtained by drawing samples $x_{0 \mid t}^{(1)},\ldots, x_{0 \mid t}^{(m)}\sim p_\theta(x_0 \mid x_t)$ via the unconditional diffusion model, and forming the estimator 
\begin{equation}
\hat{p}_{\theta}(y \mid x_t) = \frac{1}{m} \sum_{i=1}^m p(y \mid x_{0\mid t}^{(i)}).
\label{Eqn:MCCondLikeApprox}
\end{equation}
Unlike the approximation $\tilde{p}(y \mid x_t)$ in \eqref{Eqn:PointApprox}, $\hat{p}$ is an unbiased and consistent estimator of $p_\theta(y \mid x_t)$, and straightforward bounds on the Monte Carlo approximation error allow $m$ to be chosen to achieve desired accuracy.

This comes at significant cost: each sample from $p_\theta(x_0\mid x_t)$ requires running the reverse process, giving $O(mT^2)$ neural function evaluations per guided particle. The required $m$ also depends on $p(y\mid x_0)$; for indicator likelihoods with small $p_{A_y} := p_\theta(x_0\in A_y)$, naive MC requires $m \gtrsim 1/(p_{A_y}\delta\epsilon^2)$ to achieve relative accuracy $\epsilon$ with probability $\delta$. Section~\ref{sec:mlmc} addresses this with a more efficient estimator.

\citet{lgd} also employ Monte Carlo integration to approximate $p_\theta(y\mid x_t)$, but do not sample from 
 $p_\theta(x_0 \mid x_t)$ via the reverse process, instead using a Gaussian approximation centered at 
$\hat{x}_\theta(x_t)$. This approximation, while lower cost, fails to capture the multimodality of $p_\theta(x_0 \mid x_t)$ discussed above, resulting in a biased 
estimator. In contrast, using \eqref{Eqn:MCCondLikeApprox} yields an unbiased estimate of $p_\theta(y \mid x_t)$, at the cost of additional computation. In Section~\ref{sec:mlmc}, we show how to mitigate this cost through MLMC, albeit at the expense of allowing a small amount of bias. 

\subsection{Sequential Monte Carlo Sampler}

We leverage $\hat{p}_\theta(y \mid x_t)$ to construct an SMC sampler targeting the conditional distribution $p_\theta(x_0 \mid  y)$. Particles are propagated by a proposal distribution $r_t(x_{t-1} \mid x_t, y)$ and reweighted to correct for discrepancies between the proposal distribution and the true reverse conditional process. The resulting SMC algorithm is initialized by drawing particles $\{x^{(i)}_T\}_{i=1}^N \sim \mathcal{N}(\boldsymbol{0}, \I)$. Then for each time step $t$, the sampler performs two operations:
\begin{enumerate}
    \item \textbf{Proposal}: Each particle $x_{t+1}^{(i)}$ from the previous step is propagated using a proposal distribution $r_t(x_t \mid  x_{t+1}^{(i)}, y)$.
    \item \textbf{Weight \& Resample}: Compute an importance weight for each particle by calculating the approximation $\hat{p}_\theta(y\mid x_t)$ and forming
    \[w_t^{(i)} = \frac{\hat{p}_\theta(y\mid x_t) p_\theta(x_t\mid x_{t+1}^{(i)})}{\hat{p}_\theta(y \mid x_{t+1})r_t(x_t\mid x_{t+1}, y)}. \] 
    Particles are then resampled with probabilities proportional to their weights. 
\end{enumerate}
The approximation $\hat{p}_\theta(y\mid x_t)$ also implies an approximate $\hat{\nabla}_{x_t} p_\theta(y\mid x_t)$ suitable for use in \eqref{Eqn:ApproxCondScore}; however, calculation of $\hat{\nabla}_{x_t} p_\theta(y\mid x_t)$ requires computing pathwise derivatives through the stochastic reverse process, which is numerically unstable and prohibitively memory-intensive. We thus use the estimate $\hat{p}_\theta(y\mid x_t)$ directly within a weighting and resampling scheme, avoiding the need to modify the proposal distribution via unstable score approximations.
 
Because $\hat{p}_\theta(y\mid x_t)$ is an unbiased estimate of $p_\theta(y\mid x_t)$, marginally the particles follow the distribution of the conditional reverse process $p_\theta(x_t \mid x_{t+1}, y)$ (see Appendix~\ref{app:unbiased}). This contrasts with the Twisted Diffusion Sampler (TDS) \citep{wuTDS}, which also employs an SMC framework, with importance weights constructed using the point estimate $p(y\mid \hat{x}_\theta(x_t,t))$ discussed in Section~\ref{sec:marg_likes}. The resulting marginal distributions of the TDS particles can diverge from the true conditionals $p_\theta(x_t \mid x_{t+1},y)$ especially when $p_\theta(x_0 \mid x_t)$ is multimodal. This in turn can lead to extreme weight variability at the final resampling step; \citet{wuTDS} report this in the form of Effective Sample Size (ESS) collapse on MNIST and CIFAR-10. In contrast, our method maintains a stable ESS (Figure~\ref{fig:cifar10_ess_resample}).

\subsubsection{MLMC for Marginal Likelihood}\label{sec:mlmc}
To address the computational cost of sampling from $p_\theta(x_0 \mid x_t)$, we introduce a multilevel Monte Carlo (MLMC) approach \citep{Giles_2008, blanchetglynn2019}; MLMC has also recently been applied to diffusion modeling by \citet{hajiali2025}. We first define a hierarchy of discretized reverse processes $\{p_{\theta}^{(\ell)}(\cdot \mid x_t)\}_{\ell=0}^L$ with step sizes $h_\ell = h_0 M^{-\ell}$, where $h_0$ is the base step size and $M$ is an integer refinement factor \citep{Giles_2008}. Alternatively, the number of reverse steps $T_0$ at level $0$ and the refinement factor $M$ imply the number of levels $L = \log_M \left(T/T_0\right)$. Simulating the reverse process at a coarser level $\ell' < \ell$ requires fewer neural function evaluations (NFEs); however, the resulting samples $x_0^{(\ell')} \sim p_{\theta}^{(\ell')}(\cdot \mid x_t)$ are drawn from a different marginal distribution than the target $p_\theta(\cdot \mid x_t)$ due to the introduction of discretization bias. MLMC corrects for this bias by decomposing $\E_{x_0 \sim p_\theta(x_0\mid x_t)}[p(y\mid x_0)]$ into a telescoping sum of expected differences between successive discretization levels:
\begin{align}
\E[p(y\mid x_0^{(L)})] &= \E[p(y\mid x_0^{(0)})] 
+ \sum_{\ell=1}^L \E\big[ p(y\mid x_0^{(\ell)}) - p(y\mid x_0^{(\ell-1)}) \big].
\label{eq:mlmc_telescoping}
\end{align}
The MLMC estimator leverages computationally cheap samples obtained at large step-sizes, while reserving the use of fewer, more expensive, finer step-size simulations to estimate incremental corrections.
The first term is estimated using $N_0$ samples from the base (coarsest) model, with subsequent terms estimated using $N_\ell$ coupled pairs of samples to approximate the expected difference between discretization levels. The computational efficiency of MLMC relies on the variance of these corrections $V_\ell = \mathbb{V}[p(y\mid x_0^{(\ell)}) - p(y\mid x_0^{(\ell-1)})]$ decaying rapidly as $\ell$ increases. Specifically, if $V_\ell \propto M^{-\beta \ell}$ for $\beta > 0$, the number of samples $N_\ell$ required to estimate the $\ell$-th correction term decreases with level $\ell$, reducing the total computational cost.

To satisfy the variance decay condition $V_\ell\propto M^{-\beta\ell}$, we employ the synchronous coupling strategy of \citet{hajiali2025}. Coarse and fine trajectories share a common Brownian motion realization, with coarse noise increments constructed as variance-preserving weighted sums of the corresponding fine increments. The correlation between coarse and fine trajectories ensures that the expected squared difference $\E\left[\|x_0^{(\ell)} - x_0^{(\ell - 1)}  \|^2\right]$ decreases as $h_\ell \to 0$.

\begin{table}[htbp]
\centering
\caption{Empirical values of $\hat\beta$ obtained for MLMC across three
conditional-generation problems. $\hat\beta$ is estimated
from the slope of $\log V_\ell$ vs level $\ell$ over a dyadic stride ladder
$[8, 4, 2, 1]$ ($L=3$).}
\vspace{4pt}
\label{tab:beta_cross_domain}
\begin{tabular}{llrr}
\toprule
Domain & Task & $t_{\text{start}}$ & $\hat\beta$ \\
\midrule
ImageNet $256{\times}256$ & class guidance & 68 & 0.566 \\
\midrule
Stable Diffusion v1.5 & style transfer & 68 & 0.810 \\
Stable Diffusion v1.5 & style transfer & 76 & 0.858 \\
\midrule
CelebA-HQ $256{\times}256$ & age+gender & 20 & 1.234 \\
CelebA-HQ $256{\times}256$ & age+gender & 52 & 1.157 \\
\bottomrule
\end{tabular}
\end{table}

Table~\ref{tab:beta_cross_domain} shows empirically obtained values of $\hat{\beta}$ for three conditional generation tasks.
On CelebA-HQ, $\hat{\beta} > 1$ satisfies the optimality condition of \citet{Giles_2008} at both low and moderate $t$. Results for Stable Diffusion confirm that the variance reduction achieved by MLMC extends to latent-space diffusions used in high-dimensional image generation (full $\hat{\beta}$ curves shown in Appendix~\ref{app:beta_calib}).

Note that the MLMC estimator \eqref{eq:mlmc_telescoping} is a sum of estimated signed correction terms; hence if the associated Monte Carlo error is sufficiently large, $\hat{p}_\theta(y \mid x_t)$ can become negative. As $\hat{p}_\theta(y \mid x_t)$ is used to  form resampling weights, which must be nonnegative, we clamp the MLMC estimator between $0$ and $1$. This truncation creates a small bias in the estimator \eqref{eq:mlmc_telescoping}, of order the probability of negativity (which can be made arbitrarily small by increasing $N_\ell$). Thus using MLMC to compute \eqref{Eqn:MCCondLikeApprox} for computational efficiency makes the exact sampling argument in (Appendix~\ref{app:unbiased}) hold only approximately. Empirically, we find that   clamping occurs rarely (the probability of negativity is low) for chosen values of $N_{\ell}$, making the bias introduced  negligible (see Appendix~\ref{app:mlmc_clipping}).

\subsubsection{Proposal Distribution}\label{sec:smc_proposal}

The efficiency of our algorithm depends on whether the proposal places particles in high-density regions of $p_\theta(x_t\mid y)$. The unconditional kernel $r_t = p_\theta(x_t\mid x_{t+1})$ suffices in high-overlap settings (e.g. CIFAR-10), where $p_\theta(x_t\mid y)$ covers a substantial fraction of the unconditional mass. In low-overlap settings (e.g. ImageNet), the unconditional kernel rarely places particles in the target support, causing particle collapse. Our framework accommodates targeted heuristics (DPS, TFG, LGD) as proposals; reweighting then corrects their bias.

\subsubsection{Schedule-based Adaptive Resampling}\label{sec:adap_resamp}
It is common in SMC 
to reduce Monte Carlo variance by 
resampling only when the ESS drops below a predefined threshold \citep{smc_samplers}.  
However, such adaptive resampling requires tracking the ESS, and thus calculating the weights, at each time step. In our setting, calculating the importance weights $w_t^{(i)}$ requires evaluating the MLMC estimator $\hat{p}_\theta(y \mid x_t)$. Performing this evaluation at every timestep $t \in \{1, \dots, T\}$ would be prohibitively expensive, dominating the cost of the algorithm.

Instead, we employ a fixed resampling schedule defined by a subset of timesteps $\T \subset \{1, \ldots, T\}$. For a resampling step $\tau_k \in \mathcal{T}$, the sequential importance weight accumulates the discrepancies in both the likelihood and the transitions since the previous resampling step $\tau_{k-1}$:
\[ w_{\tau_k}^{(i)} = \frac{\hat{p}_\theta(y \mid x_{\tau_k}^{(i)})}{\hat{p}_\theta(y \mid x_{\tau_{k-1}}^{(i)})} \prod_{t=\tau_k}^{\tau_{k-1}-1} \frac{p_\theta(x_t \mid x_{t+1}^{(i)})}{r_{t+1}(x_t \mid x_{t+1}^{(i)}, y)}. \]
Thus we need only estimate the marginal likelihood at the resampling steps. When using the unconditional reverse process as the proposal distribution $r_t(x_t \mid  x_{t+1}^{(i)},y) = p_\theta(x_t \mid x_{t+1}^{(i)})$, the weight simplifies to the ratio of the estimated marginal likelihoods. 

The choice of $\T$ is motivated by signal-to-noise dynamics. For $t\approx T$, $\bar\alpha_t\approx 0$ implies $p_\theta(y\mid x_t)\approx p_\theta(y)$, so weights are approximately uniform and resampling provides little benefit. As $t\to 0$ the likelihood becomes informative and weight variance grows, risking degeneracy. We therefore concentrate $\T$ in the intermediate regime; for CIFAR-10 classification we find weight collapse most severe in $[70,20]$.

Of course, these boundaries will be problem-dependent. To estimate them, 
we recommend an adaptive-stepsize approach to monitoring the ESS of a pilot run (Appendix~\ref{app:res_ab}, Figure~\ref{fig:ess_cifar10}). Obtaining an approximation of the optimal resampling window may save significant resources during subsequent runs of the sampling algorithm by focusing the resampling steps within the optimal time interval.

\subsection{Algorithm}

\begin{algorithm}[htbp]
\small
\caption{SMC-MLMC for Training-Free Guidance}
\label{alg:smc_diffusion_mc}
\begin{algorithmic}[1]
\Require Particles $N$, schedule $\mathcal{T}$, proposal $r_t$, likelihood estimator $\hat{p}$ (MLMC)
\State \textbf{Init:} Sample $\{x_T^{(i)}\}_{i=1}^N \sim \mathcal{N}(0, I)$ and set prior likelihoods $\hat{L}^{(i)} \leftarrow 1$.
\For{$t = T$ \textbf{down to} $1$}
    \State \textbf{Propagate:} Sample $x_{t-1}^{(i)} \sim r_t(x_{t-1} \mid x_t^{(i)}, y)$ for all $i \in \{1,\dots,N\}$.
    \If{$t \in \mathcal{T}$}
        \State \textbf{Weight:} Compute $\hat{L}_{new}^{(i)} = \hat{p}(y \mid x_{t-1}^{(i)})$.
        \State \quad\quad\quad Calculate importance weights $w^{(i)}$.
        \State \textbf{Resample:} Sample $\{x_{t-1}^{(i)}\}_{i=1}^N$ with replacement proportional to $\{w^{(i)}\}$.
        \State \textbf{Update:} Set $\hat{L}^{(i)} \leftarrow \hat{L}_{new}^{(i)}$.
    \EndIf
\EndFor
\State \Return $\{x_0^{(i)}\}_{i=1}^N$
\end{algorithmic}
\end{algorithm}
The computational complexity of our algorithm is $O(N(T + |\mathcal{T}|ML))$, which can be decomposed into the cost of the proposal step $O(NT)$ and the reweighting $O(|\mathcal{T}|NML)$. The parameters $|\mathcal{T}|$ and $L$ control the accuracy-cost tradeoff, allowing practitioners to tune the algorithm to their requirements.

\paragraph{Hyperparameter Selection} 
Practical application of this framework requires selection of $N$, $\mathcal{T}$, $L$, and $N_\ell$. $N$ can be chosen flexibly based on compute budget. Accuracy generally increases with $N$ in classifier-guidance settings (Appendix~\ref{app:res_ab}); on CelebA-HQ joint-attribute guidance we observe non-monotonic behavior for some proposals (Appendix~\ref{app:celeba_n32}). The resampling schedule $\mathcal{T}$ is identified via a single ESS pilot run, as in Section~\ref{sec:adap_resamp}. The MLMC level count $L$ and per-level samples $N_\ell$ are determined by $\hat{\beta}$ calibration (Section~\ref{sec:mlmc}). Together, these require one ESS pilot and one $\hat{\beta}$ calibration per domain, avoiding the per-class hyperparameter tuning of \citet{dps} and \citet{tfg}.

Pure heuristic methods like DPS, which operate on single trajectories without resampling, offer no mechanism for trading increased computation for improved accuracy analogous to our ($N$,$\mathcal{T}$).
While TDS employs SMC and can eventually benefit from increased $N$, its reliance on biased likelihood approximations can lead to large distribution changes (and thus ESS collapse) along the path, resulting in poor sample complexity. 
In contrast, the weights in SMC-MLMC-DM target the true conditional distributions $p_\theta(x_t \mid y)$, ensuring that the resampling schedule provides control over the magnitude of the distribution change at each step, which in turn controls the sample complexity of the algorithm \citep{marion2024,Marion:2018b}. 

\section{Applications/Results}
\label{sec:results}
 We evaluate SMC-MLMC-DM on three tasks exercising different regimes. Section~\ref{sec:results-classifier} covers CIFAR-10 (high-overlap, unconditional proposal) and ImageNet (low-overlap, heuristic proposal); Section~\ref{sec:results-celeba} covers CelebA-HQ joint-attribute guidance, demonstrating modularity across DPS, LGD, and TFG proposals.
 
 For practical deployment, the relevant metrics are both accuracy and cost-per-success: the expected computational cost to generate at least one valid conditional sample. We define validity as the fraction of batches that produce at least one valid sample. For example, our approach achieves a $58.3\%$ success rate on ImageNet versus TFG's $20.2\%$, translating to a \textbf{1.5$\times$ cost-per-success advantage} despite higher per-run cost.

\subsection{Classifier Guidance: CIFAR-10 and ImageNet}\label{sec:results-classifier}

We evaluate on two label-guidance tasks. CIFAR-10 \citep{cifar10} (10 classes, $32{\times}32$, CIFAR10-DDPM \citep{iddpm}) is a high-overlap setting where the unconditional reverse kernel suffices as a proposal. ImageNet $256{\times}256$ (Imagenet-DDPM \citep{dhariwal2021}) is a low-overlap setting in which the target conditional support has vanishingly small volume; we  use TFG-1 \citep{tfg} as the proposal, with SMC-MLMC-DM correcting its bias.

\paragraph{Configuration.} Both tasks use $T=100$ DDIM steps. CIFAR-10: $N=16$, $r_t = p_\theta(x_t\mid x_{t+1})$, MLMC with $L=3$, $N_\ell=\{5,2,1\}$, $\mathcal{T}=\{60,50,40,30\}$. ImageNet: $N=4$, $r_t$ from TFG-1, MLMC with $L=2$, $N_\ell=\{5,1\}$, $\mathcal{T}=\{30\}$; TFG hyperparameters follow \citet{tfg}\footnote{Results shown use the 
publicly released code from 
\cite{tfg} 
with parameter settings 
reported there. 
Attempts to reproduce their results using their code yielded lower accuracy than originally reported;  this could be due to  differences in hardware or library versions. (Public comments on the Github package indicate 
other users have 
also encountered this; we contacted 
the authors without response.)
As we consider the improvement of SMC-MLMC-DM with TFG proposal over uncorrected TFG, this lower accuracy affects both and is unlikely to impact  the relative improvement.}. 
ImageNet is evaluated on classes 111, 130, 207, 222, 333, 444 (the first four following \citet{tfg}, the additional two selected at random).

\begin{table}[htbp]
    \centering
    \small
    \caption{Classifier guidance results on CIFAR-10 (10 classes) and ImageNet $256{\times}256$ (6 classes: 111, 130, 207, 222, 333, 444). Cost/Success: estimated time to generate one valid sample with probability $\geq 0.95$.}
    \label{tab:classifier_guidance}
    \begin{tabular}{llccc}
        \toprule
        \textbf{Dataset} & \textbf{Method} & \textbf{Accuracy} & \textbf{FID} & \textbf{Cost/Success (s)} \\
        \midrule
        \multirow{3}{*}{CIFAR-10}
            & TFG-1 & 52.0\% & 91.7 & 68.5 \\
            & TFG-4 & 77.1\% & 73.9 & 156.3 \\
            & \textbf{SMC-MLMC-DM} & \textbf{95.6\%} & \textbf{46.3} & \textbf{23.4} \\
        \midrule
        \multirow{2}{*}{ImageNet}
            & TFG-1 & 20.2\% & 231.9 & 230.0 \\
            & \textbf{SMC-MLMC-DM} & \textbf{58.3\%} & \textbf{182.3} & \textbf{141.7} \\
        \bottomrule
    \end{tabular}
\end{table}

\paragraph{Results.} Table~\ref{tab:classifier_guidance} reports aggregate performance. On CIFAR-10, SMC-MLMC-DM achieves $95.6\%$ Top-1 accuracy, $>18$ points above TFG-4, with a substantially lower FID, demonstrating that the unconditional reverse kernel is an effective proposal in high-overlap settings. On ImageNet, SMC-MLMC-DM nearly triples TFG's success rate  using only $N=4$ particles and a single resampling step. Despite higher per-run cost, both tasks see lower cost-per-success than baselines. The advantage scales with task difficulty: on Class 333 (Hamster), where TFG attains only $8\%$ success, SMC-MLMC-DM reaches a $2.5\times$ efficiency gain (per-class breakdown in Appendix~\ref{app:res_ab}). FID drops on ImageNet from $231.9$ to $182.3$, indicating improved sample quality alongside accuracy gains.

\paragraph{Matched Compute Experiments.} A single run of SMC-MLMC-DM requires significantly more compute than a single run of TFG. While the cost/success metric in Table~\ref{tab:classifier_guidance} corrects for runtime of a single instance, it does not take into account the fact that our algorithm takes advantage of parallelization across particles. To provide compute-adjusted comparisons, we provide experiments with matched wall-clock and matched VRAM usage. Both methods are scored by the same rule: a run succeeds if at least one output is classified as the target. K is the ratio of one complete SMC-MLMC-DM run to one TFG-1 sample, rounded up (in TFG-1's favor). 

\begin{table}[htbp]
    \centering
    \small
    \caption{TFG-1 best-of-K versus SMC-MLMC-DM (N=4) on ImageNet classifier guidance tasks with matched wall-clock time. Margin is SMC-MLMC-DM success rate minus TFG-1.}
    \begin{tabular}{lccccc}
\hline
Class & TFG-1 & $K$ & TFG best-of-$K$ & SMC-MLMC-DM ($N=4$) & Margin \\
\hline
333 (Hamster)          & 0.080 & 8 & $0.474 \pm 0.100$ & $0.487 \pm 0.033$ & $+0.013$ \\
444 (Tandem Bicycle)   & 0.102 & 4 & $0.351 \pm 0.060$ & $0.476 \pm 0.033$ & $+0.126$ \\
222 (Kuvasz)           & 0.129 & 8 & $0.669 \pm 0.083$ & $0.630 \pm 0.048$ & $-0.039$ \\
130 (Flamingo)         & 0.200 & 6 & $0.737 \pm 0.077$ & $0.570 \pm 0.033$ & $-0.167$ \\
207 (Golden Retriever) & 0.280 & 6 & $0.875 \pm 0.058$ & $0.761 \pm 0.028$ & $-0.114$ \\
111 (Nematode)         & 0.406 & 6 & $0.957 \pm 0.031$ & $0.756 \pm 0.048$ & $-0.201$ \\
\hline
\textbf{Mean} & \textbf{0.199} & & \textbf{0.677} & \textbf{0.613} & $\mathbf{-0.064}$ \\
\hline
\end{tabular}
    \label{tab:imagenet_matched_wall_clock}
\end{table}

Table~\ref{tab:imagenet_matched_wall_clock} reports matched wall clock results. We see that the mean margin across classes is slightly negative, meaning that TFG-1 is outperforming SMC-MLMC-DM. However, we see that SMC-MLMC-DM outperforms on class 333 and 444 in both experiments, providing further evidence that SMC-MLMC-DM provides gains when the targeted class is rare or hard to hit. Matched VRAM results mirror matched wall-clock; see Appendix~\ref{app:matched_vram}.

\subsection{CelebA-HQ Joint Guidance}\label{sec:results-celeba}
We evaluated SMC-MLMC-DM on CelebA-HQ $256 \times 256$ with joint-attribute guidance across the four target combinations in $\{\text{young}, \text{old}\}\times \{\text{male}, \text{female}\}$.  The guidance signal is formed from two independent classifiers, yielding a product target in which marginal accuracy on each attribute may be traded against the joint.

We use the pretrained DDPM checkpoint \texttt{google/ddpm-ema-celebahq-256} with $T = 100$ DDIM steps ($\eta = 1.0$). For SMC-MLMC-DM we set $N = 16$ with MLMC parameters determined by the  $\hat\beta$ calibration (Appendix~\ref{app:beta_calib}). We report results for $N=32$ in Appendix~\ref{app:celeba_n32}. The Twisted Diffusion Sampler (TDS) \citep{wuTDS} uses $N=16$ with its native per-step sampling schedule. Hyperparameters of TDS, as well as TFG, DPS, and LGD when used as proposals within SMC-MLMC-DM, are set according to the corresponding papers' specifications (see Appendix~\ref{app:tds_sweep}).

\begin{table}[htbp]
    \centering
    \small
    \caption{Performance on CelebA-HQ joint-attribute guidance, averaged across four targets ($\{\text{young}, \text{old}\} \times \{\text{male}, \text{female}\}$). Per-image accuracy is the mean of gender and age accuracies by held-out evaluation classifiers; per-axis columns report marginal accuracy on each classifier. Cost/Success: estimated time to generate one valid sample with probability $\geq 0.95$.}
    \label{tab:celeba_summary}
    \begin{tabular}{lcccc}
    \toprule
    \textbf{Method} & \textbf{Accuracy} & \textbf{Age} & \textbf{Gender} & \textbf{Cost/Success (s)} \\
    \midrule
    TFG (Baseline) & 0.793 & 0.784 & 0.802 & 6.5 \\
    TDS            & 0.564 & 0.548 & 0.581 & 375.9 \\
    \midrule
    \textbf{SMC-MLMC-DM (TFG-proposal)} & \textbf{0.930} & 0.921 & 0.939 & 88.3 \\
    SMC-MLMC-DM (DPS-proposal) & 0.857 & 0.844 & 0.871 & 115.1 \\
    SMC-MLMC-DM (LGD-proposal) & 0.811 & 0.707 & 0.914 & 131.3 \\
    \bottomrule
\end{tabular}
\label{tab:div_study}
\end{table}

All three SMC-MLMC-DM variants substantially outperform TDS, the most comparable SMC-based method. This gap supports our central methodological claim: TDS uses biased point-estimates in the SMC resampling step, whereas SMC-MLMC-DM targets the true marginal likelihood. Empirically, TDS suffers particle collapse, with mean validity $0.575$, whereas SMC-MLMC-DM-DPS (which uses the same reverse process as TDS) achieves validity $1.0$ on all targets (Table~\ref{tab:celeba_n16_validity}). Against the TFG baseline, SMC-MLMC-DM (TFG) increases accuracy by 14 points at higher compute cost; the baseline's 14$\times$ cost-per-success advantage reflects its high base accuracy combined with fast inference time.

\subsection{Diversity Study}
Because TDS and SMC-MLMC-DM are both SMC-based methods, we can compare the diversity of their outputs by counting the unique ancestors of their final particle sets. In Table~\ref{tab:div_study}, we report the unique ancestors, distinct images, and distinct valid images for SMC-MLMC and TDS at a range of twist scales on the CIFAR-10 label guidance task. We see the SMC-MLMC-DM produces more distinct valid images than TDS at all twist scales by a significant margin, providing evidence that SMC-MLMC-DM improves both diversity and accuracy. Note that these metrics measure particle diversity lost by the resampling step of SMC and thus do not translate to single-trajectory methods such as TFG and DPS. To compare output diversity between our method and single-trajectory methods, we use FID, reported in Table~\ref{tab:classifier_guidance} and  Appendix~\ref{app:res_ab}.

\begin{table}[t]
\centering
\caption{Diversity study}
\begin{tabular}{lccccc}
\toprule
Method & \shortstack{Twist scale\\$\alpha$}
& \shortstack{Unique\\ancestors}
& \shortstack{Distinct\\images}
& \shortstack{Distinct-\\valid}
& Accuracy \\
\midrule
\textbf{SMC-MLMC-DM} & -- & \textbf{6.1} & \textbf{9.0} & \textbf{8.5} & \textbf{0.94} \\
TDS & 0.125 & 5.6 & 7.2 & 1.2 & 0.35 \\
TDS & 0.25  & 5.6 & 7.6 & 2.5 & 0.50 \\
TDS & 0.5   & 4.1 & 6.0 & 1.7 & 0.29 \\
TDS & 1     & 2.3 & 5.3 & 1.1 & 0.26 \\
TDS & 2     & 1.2 & 3.2 & 0.4 & 0.20 \\
\bottomrule
\end{tabular}
\end{table}

\section{Limitations and Future Work}

The MLMC variance-decay rate $\hat\beta$ is problem-dependent, ranging from $0.57$ on ImageNet to $1.23$ on CelebA-HQ. The cost-per-success advantage of SMC-MLMC-DM over baseline methods depends on baseline accuracy: on tasks where baseline methods have low success rate, SMC-MLMC-DM's higher per-run cost is offset by reduced retries; on tasks where baseline already achieves high accuracy (CelebA-HQ), SMC-MLMC-DM provides accuracy gains at compute cost rather than dominating on both metrics. Within the SMC framework, however, SMC-MLMC-DM consistently dominates the closest comparator (TDS) on both accuracy and cost-per-success across all evaluated tasks. The variance of the Monte Carlo estimator of $p_\theta(y\mid x_t)$ may scale unfavorably in some high-dimensional settings (Section~\ref{sec:mc_approx}), and we are exploring algorithmic extensions to address this; \citet{feng25} documents an analogous variance challenge in flow matching. The SMC-MLMC-DM framework is formulated here for discrete-time DDPMs; continuous-time SDE adaptation follows established MLMC techniques \citep{hajiali2025}, while flow-matching adaptation requires nontrivial modification due to deterministic ODE paths, with \citet{diamondmaps} offering one direction.

\section{Conclusion}

We have presented a principled Sequential Monte Carlo framework for training-free diffusion guidance. By replacing heuristic point estimates with unbiased Monte Carlo estimators of the marginal likelihood $p_\theta(y\mid x_t)$, our approach correctly captures the multimodality of the posterior distribution.
Computational tractability is maintained by introduction of a Multilevel Monte Carlo (MLMC) estimator and a signal-adaptive resampling schedule. Empirical evaluations across CIFAR-10, ImageNet, and CelebA-HQ show accuracy improvements of 14-38 points over heuristic baselines and 25-37 points over the closest SMC-based method (TDS), with SMC-MLMC-DM beating TDS on cost-per-success across all evaluated tasks.

\newpage 

\bibliography{papers}
\bibliographystyle{plainnat}

\newpage 

\appendix

\section{Theoretical Properties}

\subsection{Exact Sampling from Unbiased Estimation}
\label{app:unbiased}
We show that, provided a nonnegative unbiased estimator of the likelihood $p_\theta(y \mid x_t)$, our algorithm targets the correct marginal distribution $p_\theta(x_t \mid y)$. We use an extended-target argument: let $m_t(u_t\mid x_t)$ be the distribution of the random variable used to construct our estimator $\hat{h}_t(x_t, u_t).$ Because our estimator is unbiased, we have
\[\int \hat{h}_t(x_t, u_t)m_t(u_t\mid x_t)du_t = p_\theta(y\mid x_t)\]
Define the unnormalized path target
\[\tilde{\gamma}_t(x_{t:T}, u_{t:T}) =  p_\theta(x_{t:T})\hat{h}_t(x_t,u_t)\prod_{s=t}^Tm_s(u_s\mid x_s)\]
By the unbiasedness of $\hat{h}(x_t, u_t),$ we have
\[\int \tilde{\gamma}_t(x_{t:T}, u_{t:T}) du_{t:T} = p_\theta(y\mid x_t) p_\theta(x_{t:T}).\]
Thus, we target the correct (unnormalized) marginal distribution. We now show that our weight formula follows from the extended target:
\[\frac{\tilde{\gamma}_t(x_{t:T}, u_{t:T})}{\tilde{\gamma}_{t+1}(x_{t+1:T}, u_{t+1:T})r_t(x_t\mid x_{t+1}, y)m_t(u_t\mid x_t)} = \frac{\hat{h}_t(x_t, u_t) p_\theta(x_t \mid x_{t+1})}{\hat{h}_{t+1}(x_{t+1}, u_{t+1})r_t(x_t\mid x_{t+1}, y)}.\]
Therefore, an unbiased estimator of $p_\theta(y\mid x_t)$ implies that our SMC algorithm targets the correct marginal distributions.

\section{Gaussian Mixture Models}
\subsection{Forward Process Marginal Distributions}
Suppose $x \in \mathbb{R}^d$ is distributed according to a GMM with $K$ components. We have $p_0(x\mid y=k) = N(x; \mu_k, \sigma_0^2 I)$ and, assuming equal component weights, $p_0(x) = \frac{1}{K}\sum_{k=1}^K N(x;\mu_k, \sigma_0^2 I).$

Consider the forward process conditioned on $y=k$; so $x_0$ is drawn from component $k$. At time $t$ we have $x_t = \sqrt{\overline{\alpha}_t}x_0 + \epsilon_t$ where $\epsilon_t \sim N(0, (1-\overline{\alpha}_t)I)$ so 
\begin{align*}
\mu_{k,t}&:=\E[x_t \mid y=k] = \sqrt{\overline{\alpha}_t}\mu_k\\
\sigma^2_t &:=\text{Cov}[x_t \mid y=k]=\overline{\alpha}_t\sigma_0^2I + (1-\overline{\alpha}_t)I = (1 - \overline{\alpha}_t(1-\sigma_0^2))I.
\end{align*}
Then the marginal distribution of the forward process at time $t$ is also a GMM: \[q_t(x_t) = \frac{1}{K} \sum_{k=1}^K N(x_t;\mu_{k,t}, \sigma^2_t I)\]
 and the membership probabilities are straightforward to calculate:
\[\Pr(y=k | x_t) = \frac{N(x_t;\mu_{k,t}, \sigma^2_tI)}{\sum_{j=1}^K N(x_t;\mu_{j,t}, \sigma_t^2I)}= \frac{\exp\left(-\frac{\|x_t - \mu_{k,t}\|^2}{2\sigma_t^2}\right)}{\sum_{j=1}^K\exp\left(-\frac{\|x_t - \mu_{j,t}\|^2}{2\sigma_t^2}\right)}. \]

\subsection{DPS estimation bias}
When $K=2$ we can use Jensen's inequality to show that DPS overestimates $p(y \mid x_t)$. Let $\mu_1 = \mu$ and $\mu_2 = -\mu$. We can rewrite the membership probability as 
\[\Pr(y=1 \mid x_t) = \frac{1}{1 + \exp\left(\frac{-2x_t^\top\mu}{\sigma_t^2}\right)}.\]

When $\frac{2x_t^\top\mu}{\sigma_t^2} > 0$ (implying $p(y=1 \mid x_t) > 0.5$), this function is strictly concave. Thus, by Jensen's inequality we have 
\[p(y=1\mid x_t) = \E_{x_0\sim p(x_0 \mid x_t)}[p(y = 1 \mid x_0)] < p(y=1 \mid \E_{x_0 \sim p(x_0\mid x_t)}[x_0]).\]
Therefore, in regions where $p(y=1 \mid x_t) > 0.5$, the DPS estimate will overestimate the class membership probability. By the same argument, when $p(y=1 \mid x_t) < 0.5$ (or $\frac{2x_t^\top\mu}{\sigma_t^2} < 0$) the DPS estimate is strictly less than the class membership probability. Because equality holds only when $x_t^\top \mu = 0$, a hyperplane of measure zero, we have 
$$P_{x_t}\left( \tilde{p}(y \mid x_t) \neq p_\theta(y \mid x_t) \right) = 1.$$

\subsection{Gaussian Likelihoods}\label{app:gaussian_like}
Assume $K=2$ with $\mu_1 = \mu$,  $\mu_2 = -\mu$, and isotropic variance $\sigma_0^2 I$ for each mode. To ensure that the modes remain distinguishable as dimension increases, take the signal to noise ratio to be constant with $d$: $\|\mu\|^2 = c\cdot d$ for some constant $c>0$. For large $t$, we will have 
\[\hat{x}_0(x_t) = \mathbb{E}[x_0 \mid x_t] \approx 0.\]
Consider a Gaussian likelihood centered at $\mu_1$ $p(y\mid x) \propto \exp\left(-\frac{\|x-\mu\|^2}{2\omega^2} \right)$ as is typical in diffusion guidance problems. Evaluating this at the point estimate gives
\[p(y \mid \hat{x}_0(x_t)) \propto \exp\left(-\frac{cd}{2\omega^2}\right)\]

\section{Experiment Details}

\subsection{CIFAR-10 Experiment Details}

\paragraph{Diffusion Model.} We use the pretrained DDPM checkpoint from 
\texttt{google/ddpm-cifar10-32} via the HuggingFace Diffusers library. 
The model operates on $32 \times 32$ RGB images with $T=100$ diffusion steps 
(mapped to the model's 1000-step schedule via stride 10).

\paragraph{Classifier.} For guidance, we use a pretrained ResNet-34 classifier 
(\texttt{resnet34\_cifar10}) from the \texttt{timm} library. Images are normalized 
using CIFAR-10 statistics ($\mu = [0.491, 0.482, 0.447]$, $\sigma = [0.202, 0.199, 0.201]$).
For evaluation, we use a separate ConvNeXT-Tiny classifier to avoid overfitting 
to the guidance signal.

\paragraph{SMC Configuration.} We use $N=16$ particles per chain. Resampling occurs 
at timesteps $\mathcal{T} = \{60, 50, 40, 30\}$ (i.e., $|\mathcal{T}|=4$ in the 
reported experiments). Resampling is triggered when ESS falls below $0.5N$.

\paragraph{MLMC Configuration.} The base integrator uses $T_0=16$ steps. We use 
$L=3$ levels with sample counts $N_\ell = \{5, 2, 1\}$ for levels $\ell = 0, 1, 2$ 
respectively. Coupled trajectories follow Algorithm~2 of \citet{hajiali2025}.

\paragraph{Evaluation.} For each of the 10 CIFAR-10 classes, we generate 2,000 samples 
(20 parallel jobs $\times$ 100 samples each). FID is computed against class-filtered 
reference images from the CIFAR-10 test set.

\subsection{ImageNet Experiment Details}

\paragraph{Diffusion Model.} We use the pretrained 256$\times$256 unconditional ImageNet 
diffusion model from \citet{dhariwal2021}, with checkpoint 
\texttt{openai\_imagenet.pt}. We use $T=100$ inference steps (subsampled from 1000 
training steps) with DDIM sampling ($\eta=1.0$).

\paragraph{Classifier.} For guidance, we use a pretrained Vision Transformer 
(\texttt{google/vit-base-patch16-224}) from HuggingFace. For evaluation, we use a 
separate DeiT-Small model (\texttt{facebook/deit-small-patch16-224}).

\paragraph{Proposal Distribution.} Unlike CIFAR-10, the unconditional reverse kernel 
is insufficient for ImageNet due to the large state space. We therefore use TFG 
\citep{tfg} as the proposal distribution $r_t(x_{t-1}|x_t, y)$.

\paragraph{SMC Configuration.} We use $N=4$ particles per chain with resampling at 
timestep $\mathcal{T} = \{30\}$ (i.e., $|\mathcal{T}|=1$). Resampling is triggered 
when ESS falls below $N/2 = 2$.

\paragraph{MLMC Configuration.} We use $L=2$ levels with sample counts 
$N_\ell = \{5, 1\}$ for levels $\ell = 0, 1$. The base integrator uses $T_0=16$ steps 
for $t > 50$, reducing to $T_0=4$ steps for $t \leq 50$ (dynamic schedule).

\paragraph{Evaluation.} We evaluate on 6 ImageNet classes (111, 130, 207, 222, 333, 444) 
with 200 samples per class. Classes 111, 222, 333, 444 were selected to match \citet{tfg}. The additional classes were selected at random. Success rate is computed as the fraction of runs producing 
at least one correctly classified particle.

\subsection{TDS Hyperparameter Selection on CelebA-HQ}\label{app:tds_sweep}

The Twisted Diffusion Sampler (TDS) of \citet{wuTDS} does not include
CelebA-HQ experiments. Since TDS uses DPS-style guidance terms within an
SMC framework, we initially set TDS hyperparameters following the
DPS-optimal configuration for CelebA-HQ from \citet{dps}. Observing
variable performance across the four joint-attribute targets
(Table~\ref{tab:celeba_per_image_full_16}) — particularly catastrophic age-axis
failure on the old+female target — we conducted a guidance-scale sweep on
this target across $\bar{\rho} \in \{1.0, 0.5, 0.25\}$ to test whether the
failure reflects insufficient hyperparameter tuning. Results are reported
in Table~\ref{tab:tds_sweep}.

\begin{table}[htbp]
\centering
\small
\caption{TDS guidance-scale sweep on target old+female (1+0). Per-image
accuracy is computed over 7 batches $\times$ 32 particles. Validity denotes
the fraction of batches producing at least one valid sample. Lower validity
indicates particle collapse.}
\label{tab:tds_sweep}
\begin{tabular}{lcccc}
\toprule
$\bar{\rho}$ & Per-image acc. & Age & Gender & Validity \\
\midrule
1.0 (paper default) & 0.440 & 0.022 & 0.857 & 1.000 \\
0.5                 & 0.571 & 0.143 & 1.000 & 0.571 \\
0.25                & 0.509 & 0.393 & 0.625 & 0.571 \\
\bottomrule
\end{tabular}
\end{table}

The configurations exhibit a non-monotonic tradeoff between attribute axes:$\bar{\rho}=1.0$ produces strong gender accuracy (0.857) but collapsed age accuracy (0.022); $\bar{\rho}=0.5$ saturates gender (1.000) but recovers age only modestly (0.143); $\bar{\rho}=0.25$ partially recovers age (0.393) at the cost of gender (0.625). Validity drops to 0.571 for the lower guidance
scales, indicating particle collapse in roughly half of all batches. While $\bar{\rho}=0.5$ wins on per-image accuracy, it loses on validity. Because validity determines cost-per-success, which we argue is the most application-relevant metric, we use $\bar{\rho}=1.0$.

\subsection{Hardware and Software}
All experiments were run using PyTorch \citep{pytorch}. We used the HuggingFace library to implement the CIFAR-10 experiments as well as details from the code released by the authors of \citet{wuTDS}. We relied heavily on the code released in conjunction with \citet{tfg} for our ImageNet experiments. 

The CIFAR-10 experiments were run on NVIDIA RTX 5000 Ada GPUs with 32GB memory. ImageNet and CelebA-HQ experiments were run using NVIDIA H200 GPUs. We estimate 150 GPU hours for the H200s and 50 GPU hours for the RTX 5000s. 

\section{Expanded Results And Ablations}\label{app:res_ab}

\subsection{ESS Dynamics and Adaptive Resampling}
\begin{figure}[htbp]
    \centering
    \begin{subfigure}[b]{0.48\linewidth}
        \centering
        \includegraphics[width=\linewidth]{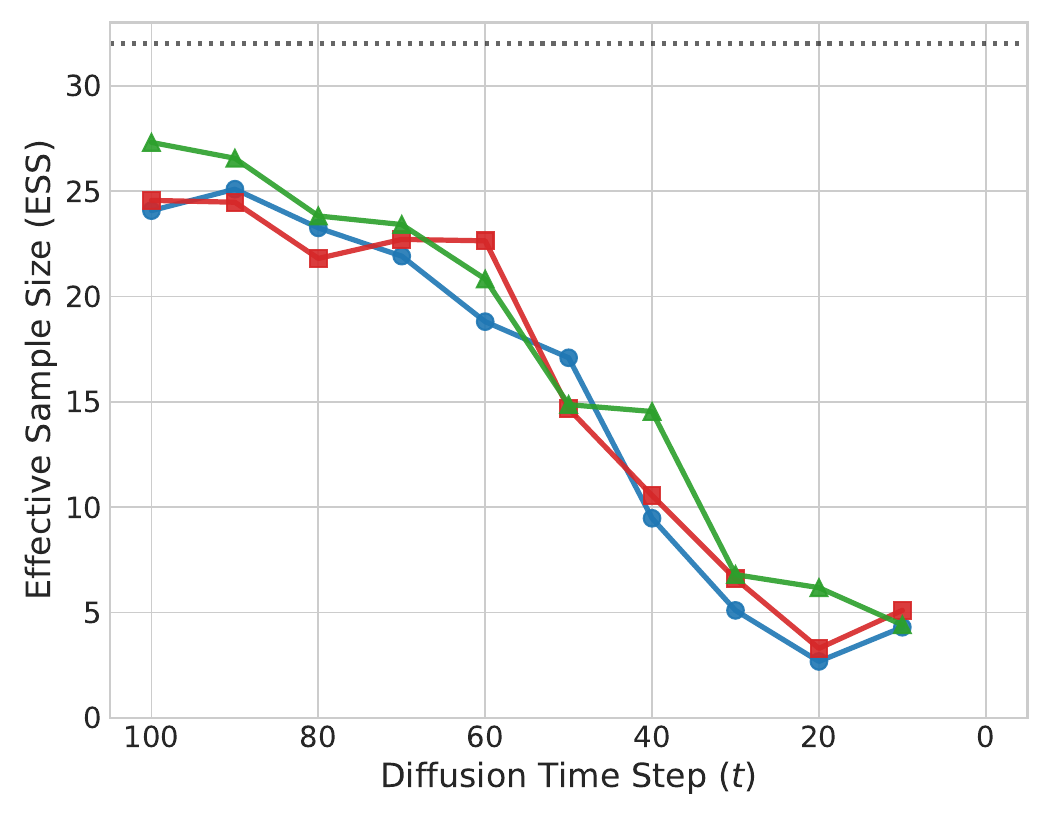}
        \caption{ESS of $32$ particles for three independent runs of CIFAR-10 classifier guidance without resampling.}
        \label{fig:cifar10_ess_exact}
    \end{subfigure}
    \hfill
    \begin{subfigure}[b]{0.48\linewidth}
        \centering
        \includegraphics[width=\linewidth]{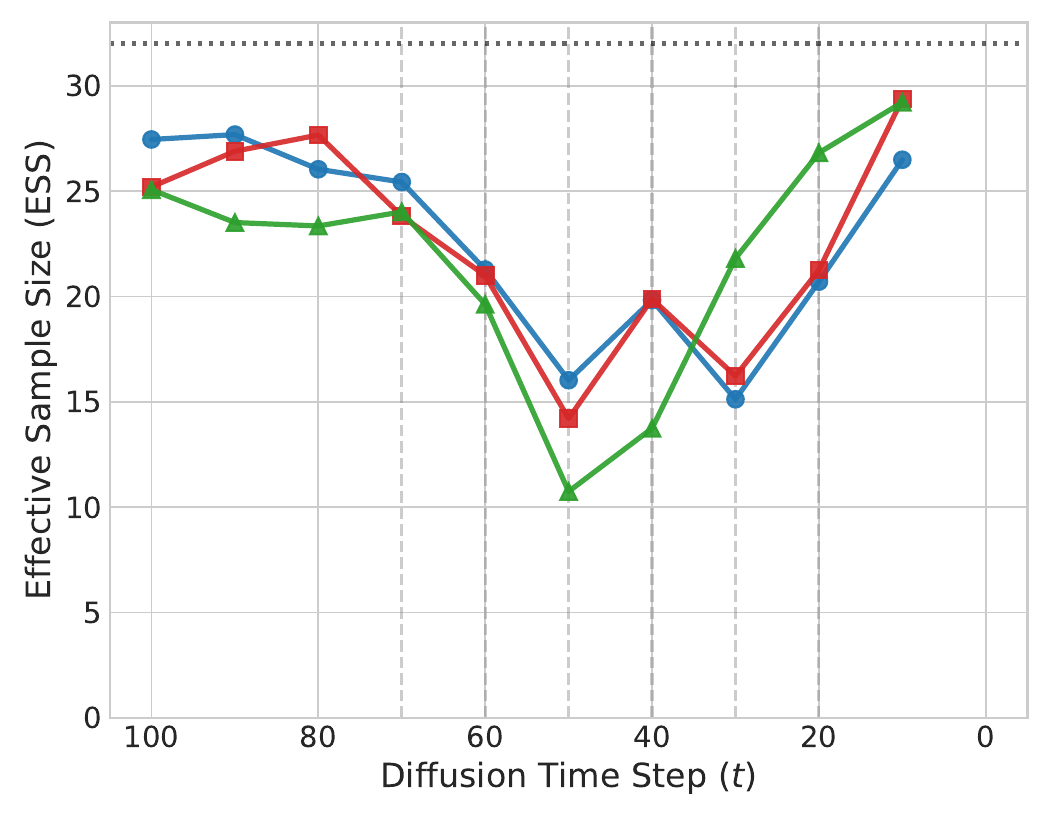}
        \caption{ESS of $32$ particles for three independent runs of CIFAR-10 classifier guidance with resampling.}
        \label{fig:cifar10_ess_resample}
    \end{subfigure}
    
    \caption{
    Six independent runs of label guidance for CIFAR-10. (a) Without resampling, ESS begins to collapse at $t=60.$ (b) Resampling is performed at $\{70,60,50,40,30\}$. The relevant window remains stable across runs, suggesting it can be identified with a single exploratory run.}
    \label{fig:ess_cifar10}
\end{figure}

We present ESS with and without resampling in Figure~\ref{fig:ess_cifar10}. Running the algorithm without resampling can identify regions of ESS collapse where resampling computation can focus.

\subsection{MLMC Clipping Experiments}\label{app:mlmc_clipping}

We compared three MLMC allocations on the same particle states at four
denoising steps, using shared sample prefixes and 256 estimates per
allocation. For each raw estimate $\hat p$, we measured the clipping
adjustment $\Delta=\operatorname{clip}(\hat p,0,1)-\hat p$.
The mean absolute adjustment and clipping frequency decreased with
larger allocations, while the signed mean varied nonmonotonically.

\begin{table}[htbp]
\centering
\small
\caption{Clipping adjustments for three MLMC sample allocations.}
\label{tab:clipping}
\begin{tabular}{lrrr}
\toprule
Allocation & Mean $\Delta$ & Mean $|\Delta|$ & Outside $[0,1]$ \\
\midrule
$(5,2,1)$  & $+0.00330$ & $0.02988$ & $30/256$ ($11.7\%$) \\
$(10,4,2)$ & $+0.00555$ & $0.01460$ & $27/256$ ($10.5\%$) \\
$(20,8,4)$ & $+0.00481$ & $0.00527$ & $13/256$ ($5.1\%$) \\
\bottomrule
\end{tabular}
\end{table}

\subsection{Matched VRAM}\label{app:matched_vram}

\begin{table}[htbp]
    \centering
    \small
    \caption{TFG-1 best-of-K versus SMC-MLMC-DM (N=4) on ImageNet classifier guidance tasks with matched VRAM. Margin is SMC-MLMC-DM success rate minus TFG-1.}
    \begin{tabular}{lccccc}
\hline
Class & TFG-1 & $K$ & TFG best-of-$K$ & SMC-MLMC-DM ($N=4$) & Margin \\
\hline
333 (Hamster)          & 0.080 & 7 & $0.434 \pm 0.094$ & $0.487 \pm 0.033$ & $+0.053$ \\
444 (Tandem Bicycle)   & 0.102 & 4 & $0.351 \pm 0.060$ & $0.476 \pm 0.033$ & $+0.126$ \\
222 (Kuvasz)           & 0.129 & 7 & $0.628 \pm 0.081$ & $0.630 \pm 0.048$ & $+0.002$ \\
130 (Flamingo)         & 0.200 & 5 & $0.679 \pm 0.074$ & $0.570 \pm 0.033$ & $-0.109$ \\
207 (Golden Retriever) & 0.280 & 5 & $0.826 \pm 0.060$ & $0.761 \pm 0.028$ & $-0.065$ \\
111 (Nematode)         & 0.406 & 5 & $0.923 \pm 0.037$ & $0.756 \pm 0.048$ & $-0.167$ \\
\hline
\textbf{Mean} & \textbf{0.199} & & \textbf{0.640} & \textbf{0.613} & $\mathbf{-0.027}$ \\
\hline
\end{tabular}
    \label{tab:imagenet_matched_vram}
\end{table}

\subsection{CIFAR-10}\label{app:res_ab_cifar10}
\begin{figure}
    \centering
    \includegraphics[width=0.5\linewidth]{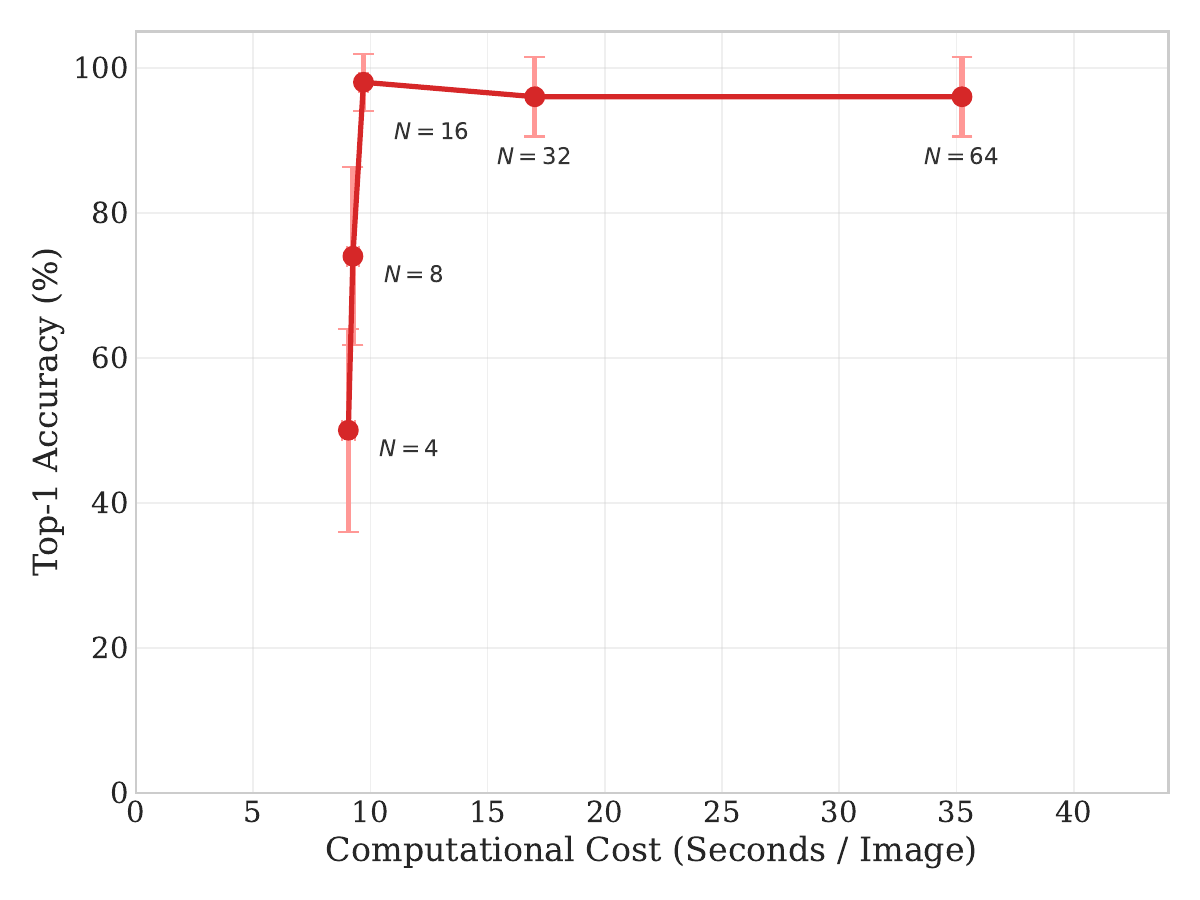}
    \caption{Computation cost versus accuracy of SMC-MLMC-DM as a function of number of particles $N$ for CIFAR-10.}
    \label{fig:ablation_particles}
\end{figure}

\paragraph{Ablation Studies} In Figure~\ref{fig:ablation_particles} we show computational cost versus accuracy for a variety of particle sizes $N$ on the CIFAR-10 label guidance task. We see that accuracy increases monotonically with $N$. Computational costs remains fixed up to $N=16$, then increases.

\paragraph{Per-Class Results} We provide the per-class accuracy and FID for SMC-MLMC-DM in Table~\ref{tab:cifar10_classwise}.
\begin{table}[htbp]
    \centering
    \caption{Per-class breakdown of SMC-MLMC-DM performance on CIFAR-10 conditional generation.}
    \label{tab:cifar10_classwise}
    \resizebox{0.5\textwidth}{!}{
    \begin{tabular}{lcc}
        \toprule
        \textbf{Target Class} & \textbf{Top-1 Accuracy} & \textbf{FID} \\
        \midrule
        Airplane & 94.9\% & 54.8 \\
        Automobile & 97.6\% & 35.0 \\
        Bird & 94.3\% & 51.0 \\
        Cat & 92.7\% & 61.1 \\
        Deer & 96.1\% & 39.0 \\
        Dog & 91.3\% & 60.2 \\
        Frog & 99.2\% & 47.6 \\
        Horse & 95.6\% & 37.3 \\
        Ship & 96.6\% & 43.2 \\
        Truck & 98.2\% & 34.1 \\
        \midrule
        \textbf{Average} & \textbf{95.6\%} & \textbf{46.3} \\
        \bottomrule
    \end{tabular}
    }
\end{table}

\begin{figure}[htbp]
    \centering 
    \begin{subfigure}[b]{0.48\textwidth}
        \centering
      \includegraphics[width=\textwidth]{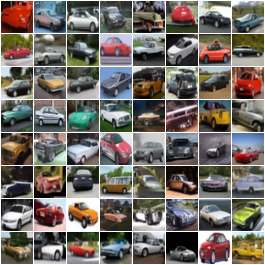}
      \caption{SMC-MLMC-DM}
    \end{subfigure}
    \hfill
    \begin{subfigure}[b]{0.48\textwidth}
        \centering
     \includegraphics[width=\textwidth]{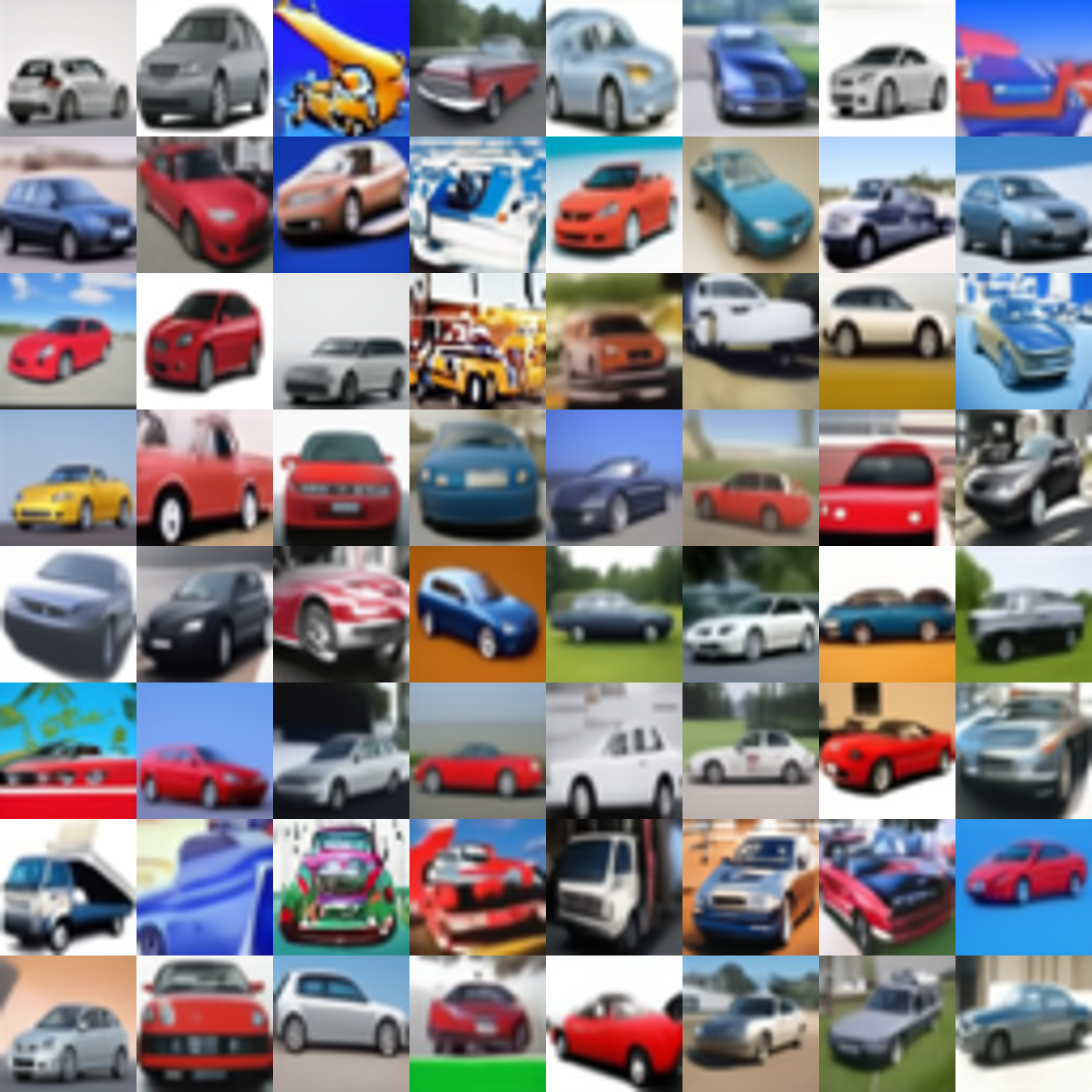}
    \caption{TFG-4}
    \end{subfigure}
    \caption{Qualitative comparison of CIFAR10 automobiles generated by (a) SMC-MLMC-DM and (b) TFG-4 \citep{tfg}.}
    \label{fig:cifar10_qual_comp}
\end{figure}

\paragraph{Qualitative Comparison} Figure~\ref{fig:cifar10_qual_comp} presents $64$ CIFAR-10 automobiles generated by SMC-MLMC-DM and TFG-4, selected at random. SMC-MLMC-DM consistently generates higher quality, more realistic images than TFG-4.

\subsection{ImageNet}
\paragraph{Particle Count Sensitivity Analysis}
\begin{table}[htbp]
    \centering
    \caption{Particle-count sensitivity on ImageNet (classes 111, 333,
    444; 10 seeds per cell, 30 runs per $N$). Standard error of the
    mean aggregates across seed and class.}
    \label{tab:imagenet_N_sweep}
    \begin{tabular}{rccc}
        \toprule
        $N$ & $n_{\text{runs}}$ & Accuracy & SEM \\
        \midrule
        4 & 30 & 53.3\% & 9.3 \\
        8 & 30 & 80.0\% & 7.4 \\
        16 & 30 & 96.7\% & 3.3 \\
        32 & 30 & 100.0\% & 0.0 \\
        \bottomrule
    \end{tabular}
\end{table}

Table~\ref{tab:imagenet_N_sweep} reports the effect of the particle
count $N$ on success rate. The sweep covers $N \in \{4, 8, 16, 32\}$
on three randomly selected ImageNet classes (111, 333, 444), with 10
seeds per $(\text{class}, N)$ configuration. Accuracy increases
monotonically with $N$, reaching $96.7\%$ at $N = 16$ and $100\%$ at
$N = 32$. This confirms that $N$ serves as a direct compute-accuracy
lever: practitioners with additional compute budget can obtain higher
accuracy without any tuning, a property that single-trajectory methods
such as DPS and TFG do not share.

\begin{figure}[htbp]
    \centering 
    \begin{subfigure}[b]{0.48\textwidth}
        \centering
      \includegraphics[width=\textwidth]{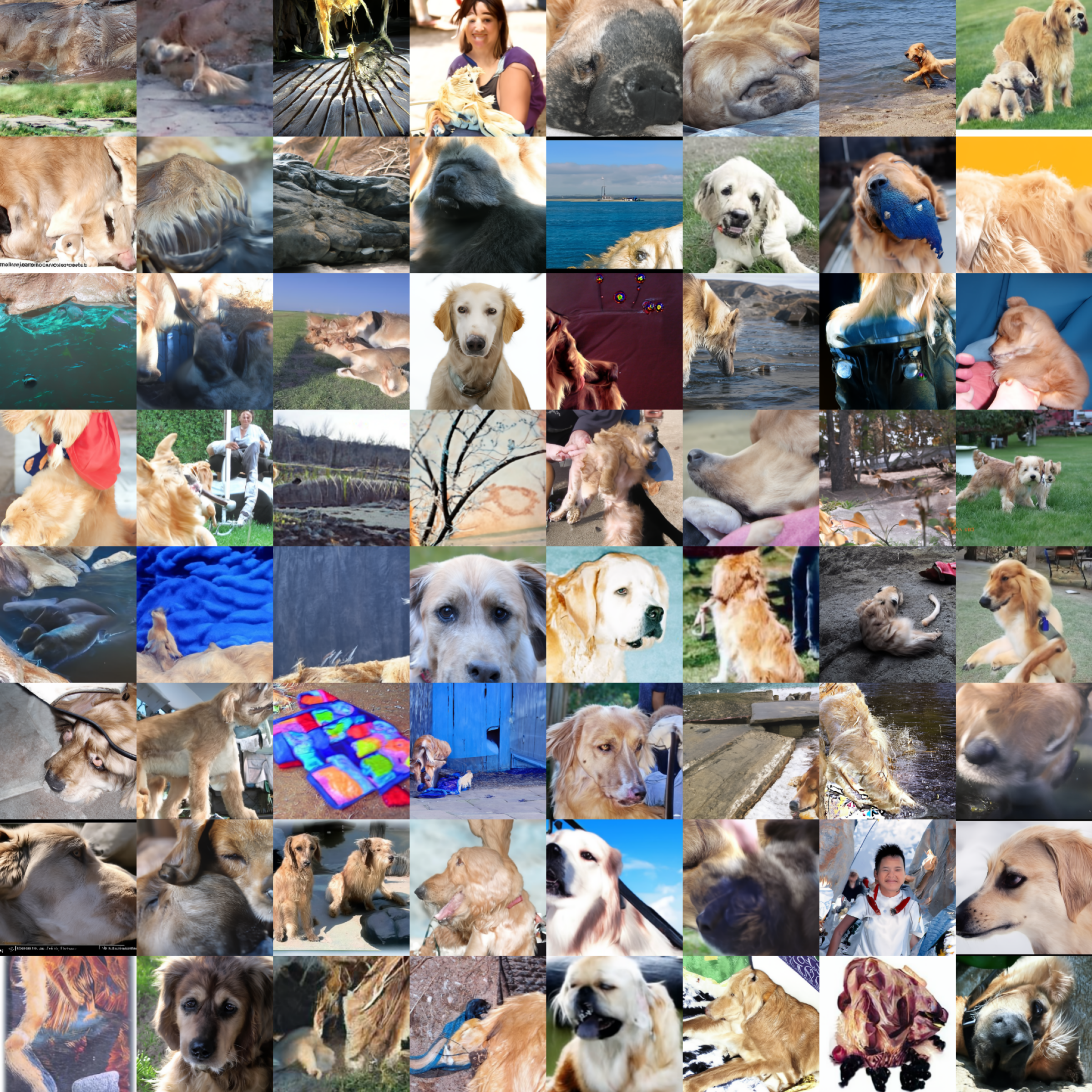}
      \caption{SMC-MLMC-DM}
    \end{subfigure}
    \hfill
    \begin{subfigure}[b]{0.48\textwidth}
        \centering
     \includegraphics[width=\textwidth]{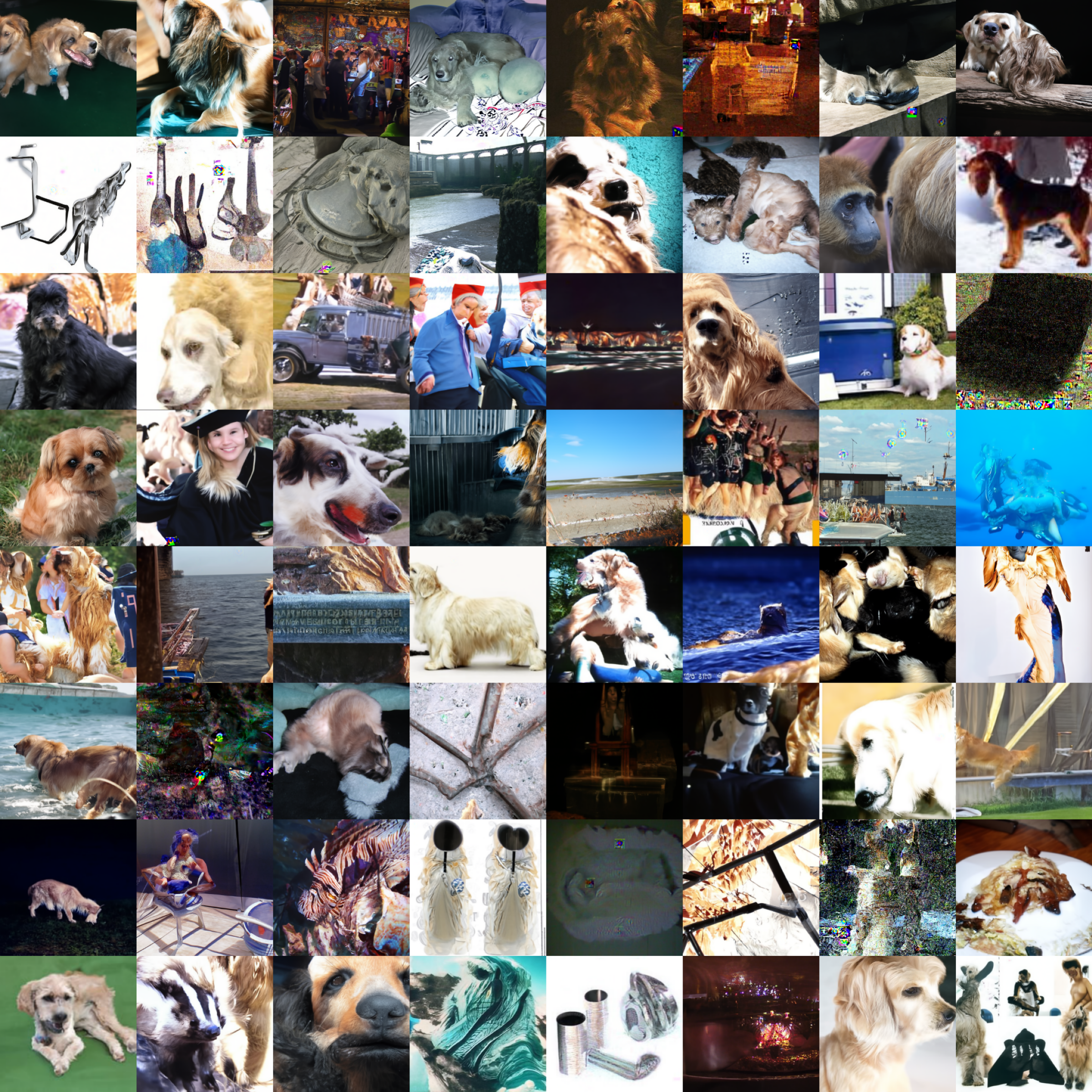}
    \caption{TFG-1}
    \end{subfigure}
    \caption{Qualitative comparison of ImageNet golden retrievers (class 207) by (a) SMC-MLMC-DM and (b) TFG-1 \citep{tfg}.}
    \label{fig:imagenet_qual_comp}
\end{figure}

\paragraph{Qualitative Comparison}
Figure~\ref{fig:imagenet_qual_comp} presents randomly selected ImageNet images generated by SMC-MLMC-DM and TFG-1 for comparison. We see a higher frequency of the target class (207, golden retriever) in (a) SMC-MLMC-DM.

\paragraph{Per-Class Results}
Table~\ref{tab:imagenet_detailed_classwise_16} provides per-class results for the ImageNet experiments.
\begin{table}[htbp]
    \centering
    \caption{Per-class SMC-MLMC-DM performance on ImageNet compared to TFG \citet{tfg}.}
    \label{tab:imagenet_detailed_classwise_16}
    \resizebox{\textwidth}{!}{
    \begin{tabular}{llcccc}
        \toprule
        \textbf{Target Class} & \textbf{Method} & \textbf{Success Rate} & \textbf{FID} $\downarrow$ & \textbf{Runtime (s)} & \textbf{Cost/Success (s)} \\
        \midrule
        \multirow{2}{*}{111 (Nematode)} & TFG (Baseline) & 41.0\% & \textbf{199.5} & \textbf{12} & \textbf{72} \\
         & \textbf{SMC-MLMC-DM} & \textbf{71.0\%} & 204.6 & 34 & 102 \\
        \midrule
        \multirow{2}{*}{130 (Flamingo)} & TFG (Baseline) & 20.0\% & 243.0 & \textbf{12} & 168 \\
         & \textbf{SMC-MLMC-DM} & \textbf{56.5\%} & \textbf{167.9} & 34 & \textbf{136} \\
        \midrule
        \multirow{2}{*}{207 (Golden Retriever)} & TFG (Baseline) & 28.0\% & 207.3 & \textbf{12} & 120 \\
         & \textbf{SMC-MLMC-DM} & \textbf{74.5\%} & \textbf{143.4} & 34 & \textbf{102} \\
        \midrule
        \multirow{2}{*}{222 (Kuvasz)} & TFG (Baseline) & 13.9\% & 232.0 & \textbf{12} & 252 \\
         & \textbf{SMC-MLMC-DM} & \textbf{51.0\%} & \textbf{199.5} & 34 & \textbf{170} \\
        \midrule
        \multirow{2}{*}{333 (Hamster)} & TFG (Baseline) & 8.0\% & 235.1 & \textbf{12} & 432 \\
         & \textbf{SMC-MLMC-DM} & \textbf{48.5\%} & \textbf{178.0} & 34 & \textbf{170} \\
        \midrule
        \multirow{2}{*}{444 (Tandem Bicycle)} & TFG (Baseline) & 10.3\% & 274.4 & \textbf{12} & 336 \\
         & \textbf{SMC-MLMC-DM} & \textbf{48.3\%} & \textbf{200.6} & 34 & \textbf{170} \\
        \bottomrule
    \end{tabular}
    }
\end{table}

\subsection{CelebA-HQ $N=16$ Additional Results}

\begin{table}[H]
      \centering
      \small
      \caption{Validity (proportion of batches that have at least one valid sample) across age and gender in CelebA-HQ joint attribute guidance (N=16).}
      \label{tab:celeba_n16_validity}
      \begin{tabular}{lccc}
      \toprule
      \textbf{Method} & \textbf{Validity} & \textbf{Age} & \textbf{Gender}  \\
      \midrule
      TFG (Baseline) & 1.000 & 1.000 & 1.000  \\
      TDS            & 0.575 & 0.563 & 0.588 \\
      \midrule
      \textbf{SMC-MLMC-DM (TFG-proposal)} & \textbf{1.000} & 1.000 & 1.000  \\
      \textbf{SMC-MLMC-DM (DPS-proposal)} & \textbf{1.000} & 1.000 & 1.000  \\
      SMC-MLMC-DM (LGD-proposal)          & 0.950 & 0.900 & 1.000  \\
      \bottomrule
      \end{tabular}
  \end{table}
  
\begin{table}[H]
\centering
\small
\caption{Per-image accuracy across four CelebA-HQ joint-attribute targets ($N=16$).}
\label{tab:celeba_per_image_full_16}
\begin{tabular}{lccccc}
\toprule
\textbf{Target} & \textbf{Baseline} & \textbf{TDS} & \textbf{TFG-prop} & \textbf{DPS-prop} & \textbf{LGD-prop} \\
\midrule
young+female (0+0) & 0.939 & 0.641 & 0.936 & 0.928 & 0.920 \\
young+male (0+1)   & 0.892 & 0.722 & 0.998 & 0.977 & 0.956 \\
old+female (1+0)   & 0.753 & 0.470 & 0.834 & 0.603 & 0.530 \\
old+male (1+1)     & 0.589 & 0.425 & 0.952 & 0.922 & 0.836 \\
\midrule
\textbf{Mean}      & 0.793 & 0.564 & \textbf{0.930} & 0.857 & 0.811 \\
\bottomrule
\end{tabular}
\end{table}

\begin{table}[H]
\centering
\small
\caption{Age-axis accuracy across four CelebA-HQ joint-attribute targets ($N=16$).}
\label{tab:celeba_age_full_16}
\begin{tabular}{lccccc}
\toprule
\textbf{Target} & \textbf{Baseline} & \textbf{TDS} & \textbf{TFG-prop} & \textbf{DPS-prop} & \textbf{LGD-prop} \\
\midrule
young+female (0+0) & 0.981 & 0.856 & 0.981 & 0.963 & 0.991 \\
young+male (0+1)   & 0.925 & 0.844 & 0.997 & 0.959 & 0.972 \\
old+female (1+0)   & 0.738 & 0.141 & 0.803 & 0.609 & 0.188 \\
old+male (1+1)     & 0.494 & 0.350 & 0.903 & 0.844 & 0.678 \\
\midrule
\textbf{Mean}      & 0.784 & 0.548 & \textbf{0.921} & 0.844 & 0.707 \\
\bottomrule
\end{tabular}
\end{table}

\begin{table}[H]
\centering
\small
\caption{Gender-axis accuracy across four CelebA-HQ joint-attribute targets ($N=16$).}
\label{tab:celeba_gender_full_16}
\begin{tabular}{lccccc}
\toprule
\textbf{Target} & \textbf{Baseline} & \textbf{TDS} & \textbf{TFG-prop} & \textbf{DPS-prop} & \textbf{LGD-prop} \\
\midrule
young+female (0+0) & 0.897 & 0.425 & 0.891 & 0.894 & 0.850 \\
young+male (0+1)   & 0.859 & 0.600 & 1.000 & 0.994 & 0.941 \\
old+female (1+0)   & 0.769 & 0.800 & 0.866 & 0.597 & 0.872 \\
old+male (1+1)     & 0.684 & 0.500 & 1.000 & 1.000 & 0.994 \\
\midrule
\textbf{Mean}      & 0.802 & 0.581 & \textbf{0.939} & 0.871 & 0.914 \\
\bottomrule
\end{tabular}
\end{table}

\subsection{CelebA-HQ $N=32$ Results}\label{app:celeba_n32}

The main results report SMC-MLMC-DM at $N=16$. Here we report results at $N=32$.

\begin{table}[H]
    \centering
    \small
    \caption{CelebA-HQ joint-attribute guidance at $N=32$, averaged across
    four targets. Per-axis columns report marginal accuracy on each classifier.
    Cost/Success: estimated time to generate one valid sample with
    probability $\geq 0.95$.}
    \label{tab:celeba_n32_summary}
    \begin{tabular}{lcccc}
    \toprule
    \textbf{Method} & \textbf{Accuracy} & \textbf{Age} & \textbf{Gender} & \textbf{Cost/Success (s)} \\
    \midrule
    TFG (Baseline) & 0.791 & 0.780 & 0.801 & 6.5 \\
    TDS            & 0.643 & 0.607 & 0.678 & 319.2 \\
    \midrule
    \textbf{SMC-MLMC-DM (TFG-proposal)} & \textbf{0.902} & 0.874 & 0.931 & 161.4 \\
    SMC-MLMC-DM (DPS-proposal) & 0.875 & 0.886 & 0.863 & 212.0 \\
    SMC-MLMC-DM (LGD-proposal) & 0.819 & 0.821 & 0.816 & 215.7 \\
    \bottomrule
\end{tabular}
\end{table}

Per-target and per-axis breakdowns at $N=32$ are reported in
Tables~\ref{tab:celeba_per_image_full_32}, \ref{tab:celeba_age_full_32}, and
\ref{tab:celeba_gender_full_32}.

\paragraph{CelebA-HQ Target Breakdowns}
Table~\ref{tab:celeba_per_image_full_32} presents accuracy across the four joint-attribute targets. Table~\ref{tab:celeba_age_full_32} presents the breakdowns for accuracy on age. Table~\ref{tab:celeba_gender_full_32} presents the breakdowns for accuracy on gender.
\begin{table}[H]
\centering
\small
\caption{Per-image accuracy across four CelebA-HQ joint-attribute targets.}
\label{tab:celeba_per_image_full_32}
\begin{tabular}{lccccc}
\toprule
\textbf{Target} & \textbf{Baseline} & \textbf{TDS} & \textbf{TFG-prop} & \textbf{DPS-prop} & \textbf{LGD-prop} \\
\midrule
young+female (0+0) & 0.942 & 0.853 & 0.938 & 0.915 & 0.833 \\
young+male (0+1)   & 0.895 & 0.708 & 0.987 & 0.989 & 0.973 \\
old+female (1+0)   & 0.754 & 0.440 & 0.797 & 0.663 & 0.580 \\
old+male (1+1)     & 0.571 & 0.571 & 0.888 & 0.931 & 0.888 \\
\midrule
\textbf{Mean}      & 0.791 & 0.643 & \textbf{0.902} & 0.875 & 0.819 \\
\bottomrule
\end{tabular}
\end{table}

\begin{table}[H]
\centering
\small
\caption{Age-axis accuracy across four CelebA-HQ joint-attribute targets,
evaluated using the held-out age classifier.}
\label{tab:celeba_age_full_32}
\begin{tabular}{lccccc}
\toprule
\textbf{Target} & \textbf{Baseline} & \textbf{TDS} & \textbf{TFG-prop} & \textbf{DPS-prop} & \textbf{LGD-prop} \\
\midrule
young+female (0+0) & 0.969 & 0.991 & 0.973 & 0.973 & 0.987 \\
young+male (0+1)   & 0.946 & 0.844 & 0.973 & 0.982 & 0.987 \\
old+female (1+0)   & 0.737 & 0.022 & 0.772 & 0.705 & 0.536 \\
old+male (1+1)     & 0.469 & 0.571 & 0.777 & 0.884 & 0.777 \\
\midrule
\textbf{Mean}      & 0.780 & 0.607 & 0.874 & \textbf{0.886} & 0.821 \\
\bottomrule
\end{tabular}
\end{table}

\begin{table}[H]
\centering
\small
\caption{Gender-axis accuracy across four CelebA-HQ joint-attribute targets,
evaluated using the held-out gender classifier.}
\label{tab:celeba_gender_full_32}
\begin{tabular}{lccccc}
\toprule
\textbf{Target} & \textbf{Baseline} & \textbf{TDS} & \textbf{TFG-prop} & \textbf{DPS-prop} & \textbf{LGD-prop} \\
\midrule
young+female (0+0) & 0.915 & 0.714 & 0.902 & 0.857 & 0.679 \\
young+male (0+1)   & 0.844 & 0.571 & 1.000 & 0.996 & 0.960 \\
old+female (1+0)   & 0.772 & 0.857 & 0.821 & 0.621 & 0.625 \\
old+male (1+1)     & 0.674 & 0.571 & 1.000 & 0.978 & 1.000 \\
\midrule
\textbf{Mean}      & 0.801 & 0.678 & \textbf{0.931} & 0.863 & 0.816 \\
\bottomrule
\end{tabular}
\end{table}

\section{MLMC Variance Decay Calibration}\label{app:beta_calib}

\subsection{Measurement Protocol}
We empirically measure the MLMC decay parameter $\hat{\beta}$ directly from coupled rollouts. For a fixed starting timestep $t_{\text{start}}$, we run the remaining $T - t_{\text{start}}$ timesteps at four dyadic stride levels $[8,4,2,1]$, giving $L=3$. At each level $\ell$, we draw coupled coarse/fine trajectory pairs using the synchronous coupling of \citet{hajiali2025} and compute the empirical variance $V_\ell$ of the increment $\log p(y \mid x_0^{(\ell)}) - \log  p(y \mid x_0^{(\ell - 1)})$. The base step count is chosen such that the stride-1 level covers the full remaining reverse process. We fit $\hat{\beta}$ as the negative slope of  $\log V_\ell$ against $\ell$ over $V_1, V_2, V_3$ as in \citet{Giles_2008}.

\subsection{Stable Diffusion}

\begin{table}[htbp]
\centering
\caption{Full $\hat\beta(t)$ calibration on Stable Diffusion v1.5 style
transfer. Six intervals across the denoising schedule, 50 trials per level.}
\label{tab:beta_sd_full}
\begin{tabular}{rrrrrrrr}
\toprule
$t_{\text{start}}$ & remaining & base & $\hat\beta$ & $V_0$ & $V_1$ & $V_2$ & $V_3$ \\
\midrule
12 & 88 & 11 & $-0.331$ & 0.0321 & 0.0577 & 0.0701 & 0.0914 \\
20 & 80 & 10 & $-0.810$ & 0.0122 & 0.0200 & 0.0377 & 0.0615 \\
36 & 64 &  8 & $+0.337$ & 0.0086 & 0.0173 & 0.0144 & 0.0108 \\
52 & 48 &  6 & $+0.505$ & 0.0137 & 0.0065 & 0.0045 & 0.0032 \\
68 & 32 &  4 & $+0.810$ & 0.0066 & 0.0029 & 0.0018 & 0.0009 \\
76 & 24 &  3 & $+0.858$ & 0.0038 & 0.0023 & 0.0019 & 0.0007 \\
\bottomrule
\end{tabular}
\end{table}

Table~\ref{tab:beta_sd_full} contains estimated $V_\ell$ and $\hat{\beta}$ for various $t_{\text{start}}$ for Stable Diffusion style guidance. High noise regimes exhibit $\hat\beta < 0$ (variance grows with level), reflecting coupling decorrelation over long rollouts; late-$t$
intervals achieve $\hat\beta > 0.8$. The sign flip at $t \approx 30$
separates the two regimes and motivates applying MLMC only in moderate to low-noise regimes on this domain. This aligns with our results in Section~\ref{sec:adap_resamp} (see Figure~\ref{fig:ess_cifar10}).

\subsection{CelebA-HQ}

\begin{table}[htbp]
\centering
\caption{Full $\hat\beta(t)$ calibration on CelebA-HQ DDPM with joint age+gender guidance.}
\label{tab:beta_celeba_full}
\begin{tabular}{rrrrrrrrrrr}
\toprule
$t_{\text{start}}$ & rem. & base & $\hat\beta$ & $V_0$ & $V_1$ & $V_2$ & $V_3$ & $\rho_1$ & $\rho_2$ & $\rho_3$ \\
\midrule
20 & 80 & 10 & $\mathbf{+1.234}$ & 12.205 & 5.685  & 1.952  & 1.027   & 0.481 & 0.148 & 0.085 \\
36 & 64 &  8 & $+0.925$          &  3.204 & 4.529  & 2.867  & 1.256   & 0.559 & 0.311 & 0.113 \\
52 & 48 &  6 & $\mathbf{+1.157}$ &  0.0124 & 0.0076 & 0.0196 & 0.00154 & 0.470 & 0.176 & 0.039 \\
68 & 32 &  4 & $+0.883$          &  0.0246 & 0.0362 & 0.0236 & 0.01064 & 0.444 & 0.145 & 0.061 \\
84 & 16 &  2 & $+0.701$          &  0.0600 & 0.0741 & 0.0355 & 0.02803 & 0.493 & 0.199 & 0.157 \\
\bottomrule
\end{tabular}
\end{table}

Table~\ref{tab:beta_celeba_full} contains the calibration results for CelebA-HQ joint age+gender guidance. The intervals ($t=20$, $t=52$) exceed Giles' classical optimality threshold $\beta > 1$. Coupling ratios $\rho_\ell = V_\ell / V_{\text{fine},\ell}$ remain below 1 at every level, confirming the synchronous coupling is active across the full $t$-range. Minor $V$-sequence non-monotonicity at intermediate levels ($V_1 > V_0$ at $t \in \{36, 68, 84\}$; $V_2 > V_1$ at $t=52$) arises because level-$\ell$ fine rollouts use $2^\ell$× more micro-steps than level-$0$ coarse rollouts, accumulating additional reverse-SDE stochasticity that the coupling cancels only partially; the $\hat\beta$ fit uses $V_1, V_2, V_3$ per MLMC convention and is unaffected.

\end{document}